\documentclass[11pt]{article}

\usepackage[preprint]{acl}

\usepackage{times}
\usepackage{latexsym}

\usepackage[T1]{fontenc}

\usepackage[utf8]{inputenc}

\usepackage{microtype}

\usepackage{inconsolata}

\usepackage{tabularx}
\usepackage{booktabs}
\usepackage{graphicx}
\usepackage{xspace}
\usepackage{xstring}
\usepackage{multirow}
\usepackage{subcaption}
\usepackage{dirtytalk}

\usepackage{longtable}
\usepackage{booktabs}
\usepackage{natbib}

\usepackage{amsmath}  %
\usepackage{amssymb}
\usepackage{siunitx}

\newcommand{\scdigits}[1]{\scalebox{0.8}{#1}}

\newcommand{\pixel}{\textsc{pixel}\xspace}

\newcommand{\pixelmeighta}{\textsc{mpixel}\textsubscript{\textsc{8l}}\xspace}
\newcommand{\pixelmeightb}{\textsc{mpixel}\textsubscript{\textsc{8s}}\xspace}
\newcommand{\pixelmsixteenbase}{\textsc{mpixel}\textsubscript{\textsc{base}}\xspace}
\newcommand{\pixelmsixteenlarge}{\textsc{mpixel}\textsubscript{\textsc{large}}\xspace}

\newcommand{\mpixel}[1][]{%
  \textsc{mpixel}%
  \ifx\relax#1\relax\else\textsubscript{\textsc{#1}}\fi
  \xspace
}
\newcommand{\pixelours}{\textsc{pixel}\textsuperscript{$\dagger$}\xspace}

\newcommand{\pixelmfour}{\textsc{pixel-m}\scdigits{4}\xspace}

\newcommand{\canines}{\textsc{canine-s}\xspace}
\newcommand{\mbert}{\textsc{mbert}\xspace}
\newcommand{\xlmr}[1][]{%
  \textsc{xlm-r}%
  \ifx\relax#1\relax\else\textsubscript{\textsc{#1}}\fi
  \xspace
}

\newcommand{\fngithub}[1]{%
  \StrBehind{#1}{https://github.com/}[\githubPathTemp]%
  \mbox{%
    \xspace
    \raisebox{-0.1ex}{\includegraphics[width=0.8em, height=0.8em]{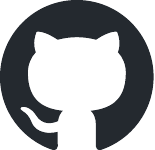}}%
    \hspace{0.2em}%
    \href{#1}{\footnotesize\texttt{\nolinkurl{\githubPathTemp}}}%
  }%
}

\title{Scaling Multilingual Pretraining of Pixel Language Models}

\author{Ilker Kesen$^{\dagger}$\thanks{Work done while at the University of Copenhagen.} \ \ Jenny Kunz$^{\ddagger}$ \ \ Kasra Malihi$^{\S}$ \ \ Desmond Elliott$^{\S}$\\
$^{\dagger}$Humboldt-Universität zu Berlin \ \
$^{\ddagger}$Linköping University \ \
$^{\S}$University of Copenhagen \\
\texttt{ilker.kesen@hu-berlin.de}}

\begin{document}
\maketitle
\begin{abstract}
Pixel language models process text as rendered images instead of a finite vocabulary, which gives them strong cross-lingual transfer abilities. 
However, not much is known about how these abilities are influenced by larger-scale pretraining.
In this paper, we train a family of \mpixel models along three axes of variation: more languages, more scripts, and larger model capacity under fixed per-language compute.
In evaluations on five tasks covering 83 languages, we find that scaling acts locally: each axis improves the added languages and scripts, while increased capacity mitigates the curse of multilinguality from pretraining on more languages.
Transfer to unseen scripts improves with multilingual pretraining, but does not further improve with scaling.
On typologically distant low-resource languages, our models outperform subword and character-level baselines across many writing systems.
We also show that larger pixel models reserve more layers for visual processing, and that cross-lingual alignment weakens as more languages are added.
Overall, our findings characterize how multilingual scaling improves pixel language models.

\end{abstract}

\section{Introduction}
\label{sec:introduction}
Pixel language models process text as rendered images rather than discrete tokens \citep{salesky-etal-2021-robust,rust2023language}. This makes them appealing for the long tail of the world's writing systems~\citep{joshi-etal-2020-state}, with previous research demonstrating strong transfer abilities to unseen scripts~\citep{kesen-etal-2025-multilingual,lotz2025overcoming}. 
However, there is an open question about whether such transfer can be improved by increasing (i) the number of writing systems, (ii) the pretraining languages, or (iii) by increasing model capacity.

\begin{figure}[!t]
    \centering
    \includegraphics[width=\columnwidth]{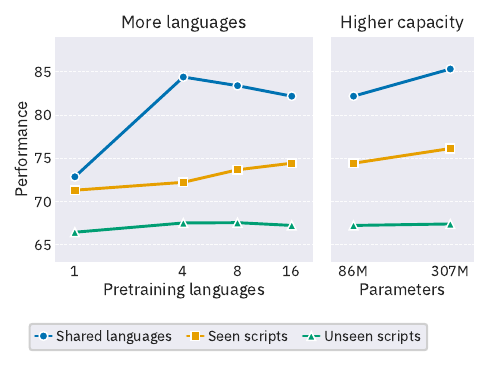}
    \caption{\mpixel performance when scaling the number of pretraining languages and model capacity. 
    Adding new languages steadily improves performance on seen scripts, albeit with a slight decrease on shared languages.
    Higher capacity improves performance on shared and seen scripts. 
    On unseen scripts, neither additional languages nor capacity yields consistent gains.}
  \label{fig:fig1}
\end{figure}

We investigate this question in a carefully controlled study that separates these axes of multilingual scaling.
Our starting point is an 86M parameter model pretrained on English, Hindi, Traditional Chinese, and Ukrainian, following \citet{kesen-etal-2025-multilingual}.
From here, we train two models to investigate the effect of increasing the number of writing systems.
The first is pretrained on eight-languages by adding four languages in the already-seen Latin script, whereas the other model adds four languages in previously unseen writing systems.
We also pretrain an \mpixel model by combining the seen and unseen scripts with four further languages, resulting in a total of sixteen pretraining languages. 
Finally, \mpixel[large] increases the model capacity to $307$M parameters.

These pretraining variations allow us to study the following research questions.
\textbf{RQ1:} Along which axis does scaling multilingual pretraining yield greater cross-lingual transfer: increasing language diversity within an already seen script or increasing script diversity?
\textbf{RQ2:} Does scaling multilinguality along both axes continue to improve cross-lingual transfer, or do gains saturate with more pretraining languages?
\textbf{RQ3:} To what extent does increased model capacity improve cross-lingual transfer?
\textbf{RQ4:} Do pixel language models suffer from the curse of multilinguality \citep{conneau-etal-2020-unsupervised} as more languages are added, and if so, does increased model capacity mitigate it?

We answer these questions with empirical experiments on five downstream tasks: sentence classification, dependency parsing, named entity recognition, natural language inference, and English-only sentence classification using the GLUE benchmark \citep{wang-etal-2018-glue}.
We evaluate on 83 languages in total, grouped into pretraining languages, unseen languages in pretraining scripts, and languages in unseen scripts.
Among these, we prioritize low-resource languages, including some that are typologically distant from the pretraining languages of all baselines.
We also analyze the models' representations by testing cross-lingual alignment with ABX discrimination tasks, and locate where scaling acts by probing linguistic information across layers at the sentence and word level. 

Our contributions are summarized as follows:

\begin{itemize}
    \item We introduce \mpixel, a family of pixel language models\footnote{Code and models: \fngithub{https://github.com/ilkerkesen/mpixel}} pretrained on up-to sixteen languages at 86M and 307M parameters.
    This includes the first pixel language models pretrained on right-to-left scripts.
    \item We show that increasing the number of pretraining languages and scripts mainly benefits the newly covered languages and scripts. %
    Transfer to unseen scripts is established by multilingual training, but does not improve with further scaling.
    \item We show that pixel language models suffer from the curse of multilinguality, and that increased capacity mitigates it.
    \item We show that \mpixel outperforms subword and character-level baselines on low-resource languages typologically distant from XLM-R pretraining languages.
    \item We find that 307M parameter \mpixel model reserves more layers for visual processing, and that cross-lingual alignment weakens as languages are added but generalizes to unseen languages through shared script.
\end{itemize}

\begin{table}[t]
\centering
\renewcommand{\arraystretch}{1.1}%
\begin{tabular}{lcccc}
\toprule
               & $\theta$ & L & Patches & Words \\ \midrule
\mpixel        & ~86M       & 16  & 120B     & 38B   \\
\mpixel[large] & 307M       & 16  & 120B     & 38B   \\ \midrule
\mpixel[4s]    & ~86M       & ~4  & ~30B     & 10B   \\
\mpixel[8l]    & ~86M       & ~8  & ~60B     & 19B   \\
\mpixel[8s]    & ~86M       & ~8  & ~60B     & 19B   \\ \bottomrule
\end{tabular}
\caption{Overview of the trained models. The \mpixel and \mpixel[large] models as the main outcomes of this paper, while the \textsc{4s}, \textsc{8l} and \textsc{8s} models are used in our control experiments. $\theta$: number of parameters. L: number of pretraining languages. Patches: number of unique image patches seen during pretraining. Words: approximate number of white-space delimited words.}
\label{tab:model-configurations}
\end{table}

\section{Pretraining Setup}
\label{sec:pretraining}

\paragraph{Models.}
We pretrain five models to support our study on scaling pixel language models.
These models vary in the number of pretraining languages, the diversity of those languages, and the number of trainable parameters, as outlined in Table~\ref{tab:model-configurations}.

\paragraph{Axes.}\hspace{-2ex}
The models are scaled along three axes: language diversity within a seen script (\mpixel[4s] $\rightarrow$ \mpixel[8l]), script diversity (\mpixel[4s] $\rightarrow$ \mpixel[8s]), and model capacity (\mpixel{} $\rightarrow$ \mpixel[large]).
Since new scripts necessarily arrive with new languages, the two data expansions are matched in language count and differ only in whether the additions introduce new scripts; contrasting them asks whether script diversity adds anything beyond language diversity (RQ1).
\mpixel{} expands along both data axes at once (RQ2), \mpixel[large] varies capacity on this data (RQ3), and evaluating every variant on the four shared pretraining languages quantifies the curse of multilinguality and whether capacity mitigates it (RQ4).

\paragraph{Languages.} We use FineWeb and multilingual FineWeb2 corpora \citep{penedo2024fineweb,penedo2025fineweb} for pretraining data.
When selecting pretraining languages, we only consider higher-resource languages, and deliberately exclude some high-resource languages for downstream task evaluations.
As mentioned earlier, we maximize the diversity in pretraining languages along two axes: language family and writing system.
For this purpose, \mpixel[8l] extends \mpixel[4s] with four Latin-script languages from diverse language families (Spanish: Romance, Danish: Germanic, Finnish: Uralic, Turkish: Turkic), varying language family while holding the writing system fixed.
\mpixel[8s] instead extends \mpixel[4s] with four languages in previously unseen scripts (Korean: Hangul, Thai: Thai, Greek: Greek, Standard Arabic: Arabic), varying both writing system and language family. 
\pixelmsixteenbase combines these extensions and adds four further languages: Vietnamese (Latin), Persian (Perso-Arabic), Bengali (Bengali), and Hebrew (Hebrew), further extending coverage along both axes.
For each language, we  an equal amount of patches, which is roughly equal to 7.5 billion image patches.
We fix the maximum number of patches per example to $512$.

\paragraph{Rendering.}
As demonstrated by \citet{lotz-etal-2023-text}, rendering strategy is a crucial design choice for pixel language models, where packing character bigrams per patch decreases input redundancy and improves performance.
We design a new \textsc{adaptive} rendering strategy for multilingual processing.
By default, we prefer grapheme-aware bigrams grouping as the original \textsc{bigrams} rendering may separate diacritics from their base characters.
For CJK languages, where wide characters make most bigrams span multiple patches, we render one character per patch, which lowers input redundancy at the same sequence length.
For Arabic and Hebrew, the \textsc{bigrams} strategy can sever ligatures, and rendering in visual order distorts word structure.
We therefore use word-level rendering for these scripts, where each word is rendered as a contiguous unit in logical order.

\paragraph{Pretraining.}
Our pretraining procedure builds on prior work.
Our model variants implement the Vision Transformer Masked Autoencoder architecture (ViT-MAE) \citep{He_2022_CVPR} by following the previous \pixel models.
The only hyperparameter we vary independently is the effective batch size, which we double for the 8-language variants and quadruple for the 16-language variants.
With 7.5B patches per language, 1M steps, and 512 patches per sample, this holds per-language exposure at ${\approx}4.3$ epochs, which a fixed batch size would otherwise halve with each doubling of the language count.
We therefore double and quadruple the batch size for the eight- and sixteen-language model variants.
As we increase the batch size, we also adjust the learning rate by following the square root scaling procedure \citep{li2024surge,krizhevsky2014weirdtrickparallelizingconvolutional}.
Unlike previous \pixel models, we select checkpoints on a held-out task:
zero-shot bitext mining on FLORES-200 \citep{nllb2022} over each variant's pretraining languages.
We exclude examples overlapping with SIB-200 \citep{adelani-etal-2024-sib} to prevent leakage.
The rest of the pretraining recipe is identical with previous \pixel models.

\section{Experimental Setup}
\label{sec:experimental-setup}

\paragraph{Baselines.}
We compare against several bidirectional encoder-based language models, including sub-word tokenizer-based \xlmr \citep{conneau-etal-2020-unsupervised} and \textsc{mbert} \citep{devlin-etal-2019-bert}, character-level hash embedding-based \textsc{canine} \citep{clark-etal-2022-canine}.
In order to eliminate confounding factors of data quality, data size, and text rendering, we pretrained our own \pixel and \pixelmfour models, which are denoted as \pixelours and \mpixel[4S].\footnote{Both models are pretrained on the same 30B-patch budget, which \pixelours spends entirely on English, while \mpixel[4S] splits the budget equally across its four languages.}

\begin{table*}[t!]
\centering
\renewcommand{\arraystretch}{1.3}%
\begin{tabular}{@{}lrrrcccccc@{}}
\toprule
\multirow{1}{*}{Model} & \multirow{1}{*}{$|\theta|$} & \multirow{1}{*}{\#L} & \multirow{1}{*}{$|X|$} & SIB-200 {\small (77)} & NER {\small (12)} & UDP {\small (30)} & XNLI {\small (15)} & GLUE {\small (1)} & \multirow{1}{*}{Avg} \\
\midrule
\canines & 127M & 104 & 1.0T & 55.8 & 70.8 & 69.7 & 70.5 & 73.0 & 68.0 \\
\mbert & 179M & 104 & - & 63.0 & \underline{81.6} & 80.0 & 74.7 & \underline{79.6} & 75.8 \\
\xlmr & 270M & 100 & 6.3T & \textbf{65.4} & \textbf{83.1} & \textbf{82.7} & \textbf{79.1} & \textbf{80.8} & \textbf{78.2} \\ \hline
\pixelours  & 86M & 1  & 0.1T  & 52.9  & 73.6  & 75.7  & 67.1     & 78.7     & 69.6 \\
\mpixel & 86M & 16 & 0.5T & 61.9 & 77.3 & 80.3 & 69.5 & 73.3 & 72.5 \\
\mpixel & 307M & 16 & 0.5T & \underline{64.4} & 80.9 & \underline{80.7} & \underline{75.0} & \underline{79.6} & \underline{76.1} \\
\bottomrule
\end{tabular}
\caption{Performance of \mpixel{} compared to subword vocabulary and character-level models. 
Each cell is the average over all evaluation languages of the best three runs per language, selected by validation performance. 
We report F1 for SIB-200 and NER, LAS for UDP, accuracy for XNLI and the average score for GLUE, over ($\cdot$) languages. 
$|\theta|$: model size. 
\#L: number of pretraining languages. 
$|X|$: pretraining size in characters for \canines, subwords for \xlmr, and image patches for the \mpixel models. 
Best results are in \textbf{bold}, second best \underline{underlined}.}
\label{tab:baselines}
\end{table*}

\paragraph{Tasks \& Benchmarks.}
We evaluate models on sentence classification on 77 languages from SIB-200 \citep{adelani-etal-2024-sib}.
We focus on (i) languages seen during pretraining; (ii) languages in seen scripts during pretraining; and (iii) languages in unseen scripts (e.g., N'ko).\footnote{We prioritize low-resource languages, i.e., classes 0--2 in the taxonomy of \citet{joshi-etal-2020-state} or absent from it. See Appendix for the list of evaluation languages.}
We also test models on dependency parsing using Universal Dependencies \citep{nivre-etal-2020-universal} under the same transfer scenarios; on named entity recognition across 12 languages\footnote{We combine data from Universal NER \citep{mayhew-etal-2024-universal,blevins-etal-2026-universal}, Naamapadam \citep{mhaske-etal-2023-naamapadam}, NER-UK \citep{chaplynskyi-romanyshyn-2024-introducing} and KLUE \citep{park2021klue}}; on natural language inference with XNLI \citep{conneau-etal-2018-xnli} (\textit{translate-train-all}); and on GLUE \citep{wang-etal-2018-glue}, which we use to investigate the \textit{curse of multilinguality} in English.

\section{Results}
\label{sec:results}

We first compare the \mpixel models against the baselines in Table~\ref{tab:baselines}.
We find that \xlmr obtains the best overall averages on all tasks, with \mpixel[large] the second-best performing model.
We note that \xlmr and \canines are pretrained on ${\sim}100$ languages, which means they have been trained on most of the languages in our seen-script and unseen-script groups.
We also note that \xlmr has seen much more data per language than \mbert, \canines, or the \mpixel models.\footnote{
For English alone, the \xlmr pretraining corpus contains ${\sim}55$B tokens.
On the other hand, our \mpixel pretraining corpus is $\sim$2.5B English whitespace separated words.}
We find that \mpixel outperforms the character-level \canines on all tasks except XNLI.
On UDP, \mpixel achieves an overall better score than \xlmr on unseen scripts (Table~\ref{tab:udp}).
However, this is largely driven by Coptic, which uses a writing system not included during \xlmr pretraining.

\begin{figure}[!t]
    \centering
    \includegraphics[width=\columnwidth]{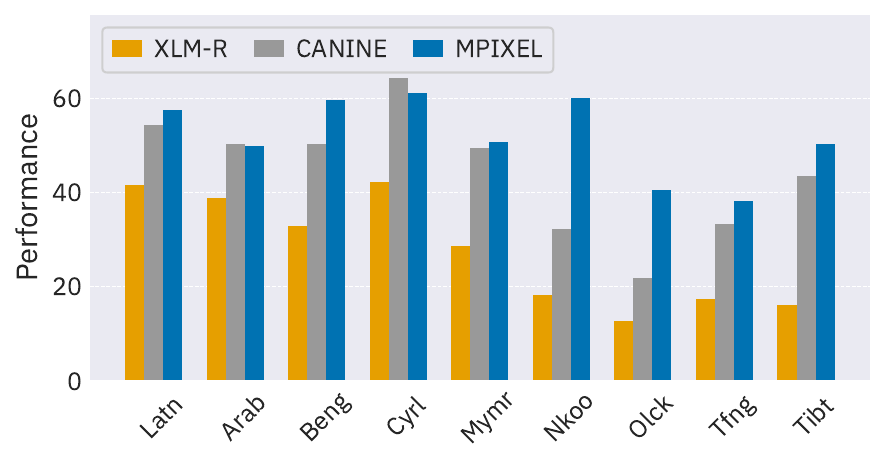}
    \caption{Performance of models on typologically distant low-resource languages in the SIB-200 benchmark.}
    \label{fig:sib200-distant-languages}
\end{figure}

\paragraph{Low-resource cross-lingual transfer.}
For a controlled evaluation of low-resource cross-lingual transfer, we compare performance on languages unseen by all models.
Within our evaluation set, we use the DistaLs toolkit~\citep{goot-etal-2025-distals} to determine which languages are typologically distant from every model's pretraining data.
Figure~\ref{fig:sib200-distant-languages} shows the SIB-200 results on these languages.
\mpixel outperforms \xlmr on all writing systems.
\mpixel also outperforms \canines on Latin and Bengali.
The vocabulary bottleneck explains this gap at two levels.
On seen scripts, the \xlmr subwords are well-trained for its pretraining languages, so they segment and represent distant languages poorly.
On the four scripts unseen by all models (N'ko, Ol Chiki, Tifinagh, Tibetan), the characters fall outside the vocabulary and are largely mapped to unknown tokens.
\xlmr performs close to random on these scripts, whereas \canines and \mpixel demonstrate their strengths.

\begin{figure*}[!t]
    \centering
    \includegraphics[width=\textwidth]{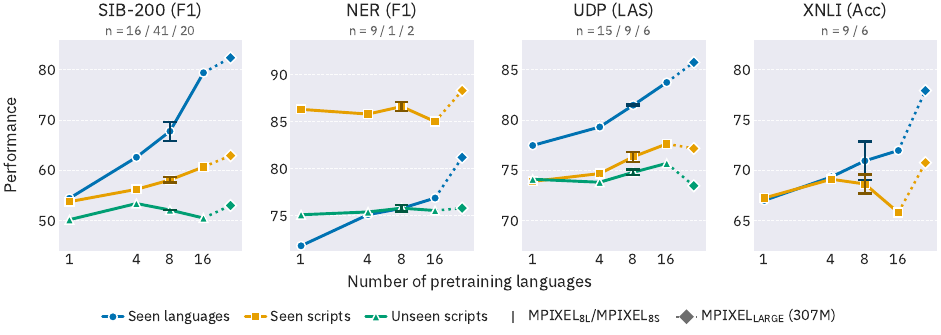}
    \caption{Downstream performance vs.\ number of pretraining languages (86M models; $x{=}1$ is English-only). Lines are evaluation tiers fixed w.r.t. \mpixel's languages and scripts; each point is a mean over that tier's items ($n$ per panel, in legend order). At $x{=}8$ the point is the mean of \mpixel[8l] and \mpixel[8s] and the bar spans them.}
    \label{fig:scaling-across-benchmarks}
\end{figure*}

\begin{figure}[!t]
    \centering
    \includegraphics[width=\columnwidth]{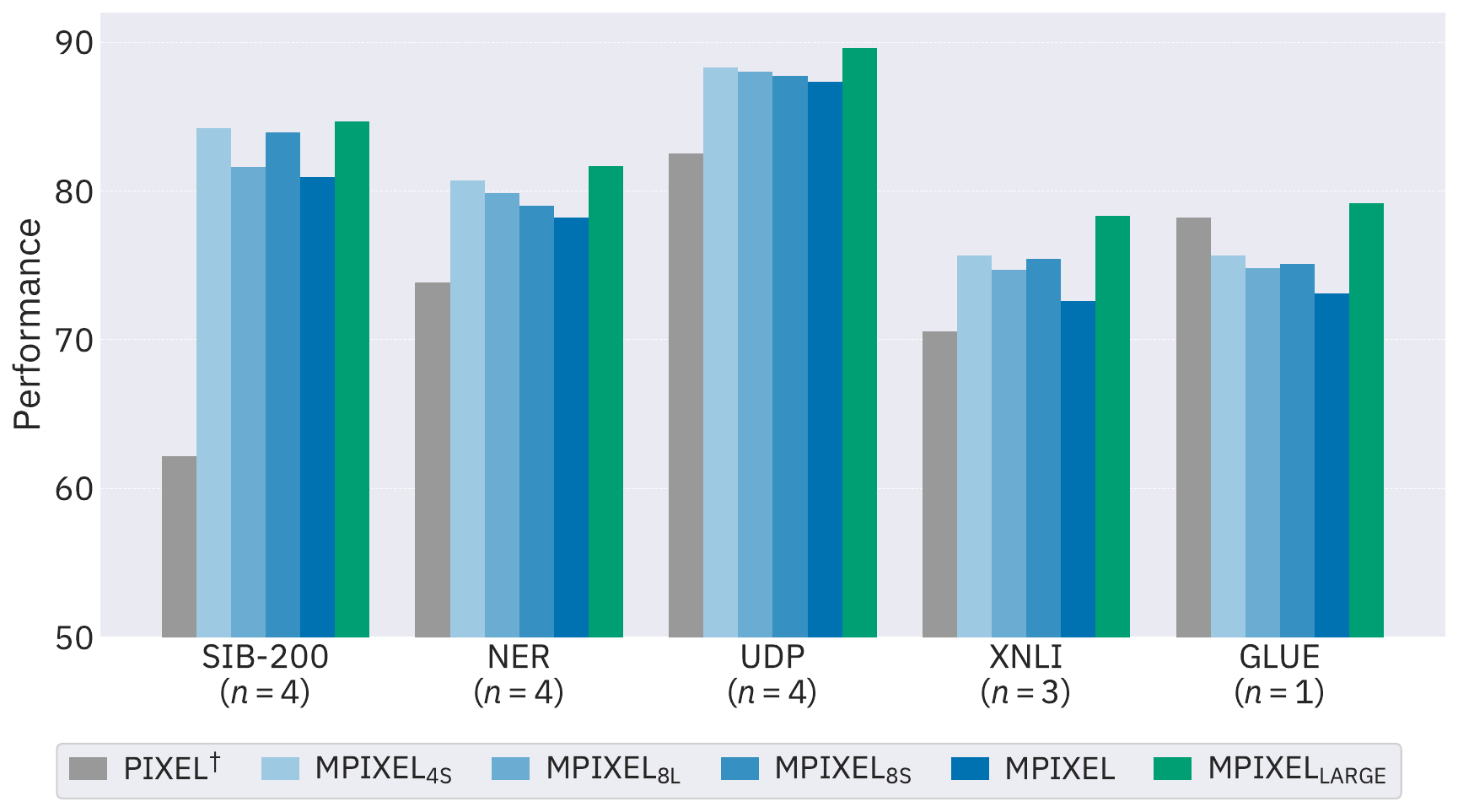}
    \caption{Performances of the \mpixel variants on the downstream benchmarks considering only four pretraining languages shared across models.}
    \label{fig:curse-of-multilinguality}
\end{figure}

\vspace{0.5em}
\par Figure~\ref{fig:scaling-across-benchmarks} provides an overview of how downstream performance changes across the language and capacity scaling settings considered in our research questions (RQ1--RQ3).

\paragraph{RQ1: Expanding language and script coverage mainly benefits the languages and scripts it adds.}
We compare \mpixel[8L] and \mpixel[8S] against \mpixel[4S], where the former adds four Latin-script languages, and the latter four languages in previously unseen scripts.
Figure~\ref{fig:scaling-across-benchmarks} shows clear improvements when pretraining on four additional languages, albeit with smaller improvements on the seen-script group for SIB-200 and UDP, and overall mixed benefits for unseen scripts.
For both of the 8-language variants, gains concentrate on the added languages and on other languages sharing their scripts: Latin for \mpixel[8L] and the newly covered scripts for \mpixel[8S] (e.g., the Arabic script; Tables~\ref{tab:sib}, \ref{tab:ner}, and \ref{tab:udp})
NER and XNLI cover too few seen-script languages for a reliable comparison.
On unseen scripts, the changes are small and mixed for both variants:
neither addition therefore yields a consistent advantage on unseen scripts.

In order to isolate the effect of multilinguality from training volume and other pretraining confounds, we train our own monolingual \pixelours with the same patch budget as \mpixel[4S].
In this matched setup, \mpixel[4S] outperforms \pixelours on unseen scripts in SIB-200 (see Table~\ref{tab:sib} in the Appendix).
The gains also extend to scripts visually unrelated to Devanagari, such as N'ko and Ethiopic, so script similarity alone cannot explain transfer.

\paragraph{RQ2: Scaling further along the language and script axes brings further improvements for seen languages and their scripts.}
If we compare the 16-language \mpixel[] against \mpixel[8L] and \mpixel[8S], the overall averages improve on SIB-200, NER and UDP, as shown in Figure~\ref{fig:scaling-across-benchmarks} (see Tables~\ref{tab:sib}, \ref{tab:ner}, and \ref{tab:udp} for detailed per-language breakdown).

The improvements concentrate on seen languages, where the added languages improve their own performance.
They partly extend to seen scripts, which improve on SIB-200 and UDP but decrease on NER and XNLI (Figure~\ref{fig:scaling-across-benchmarks}); nevertheless, NER and XNLI cover only a few seen-script languages.
On unseen scripts, the effect depends on the task.
On SIB-200 (Table~\ref{tab:sib}), \mpixel[] obtains the lowest score among the multilingual variants; on NER (Table~\ref{tab:ner}), all \mpixel[] variants stay within one point.
On UDP (Table~\ref{tab:udp}), unseen scripts instead improve with each extension, and \mpixel[] outperforms all other models, including \textsc{xlm-r}.

\paragraph{RQ3: Increased capacity only improves performance on seen languages and scripts.}
We compare \mpixel[large] (307M) against \mpixel[] (86M) pretrained on the same data.
As shown in Figure~\ref{fig:scaling-across-benchmarks}, seen languages improve on every task, with the largest gains on NER and XNLI.
Seen scripts improve on SIB-200, NER, and XNLI, and decrease slightly on UDP.
On unseen scripts, the changes are mixed: SIB-200 improves, NER stays roughly equal, and UDP decreases.
With this decrease, \mpixel[large] loses the UDP advantage on unseen scripts, falling below all smaller multilingual variants.
Even where unseen scripts improve, \mpixel[large] does not exceed \mpixel[4S] on SIB-200.
Increased capacity therefore improves performance within the pretraining scripts, while its effect on unseen scripts is inconsistent across tasks.

\paragraph{RQ4: Pixel models exhibit the curse of multilinguality, but increased capacity mitigates it.}
To measure the curse of multilinguality in pixel models, we evaluate all variants on the four shared pretraining languages (Figure~\ref{fig:curse-of-multilinguality}).
Across the multilingual variants, per-language pretraining data and compute are fixed, so performance differences cannot be explained by reduced exposure to these languages.
From \mpixel[4S] to \mpixel[], performance declines on all five benchmarks.
On GLUE, English performance drops from the monolingual \pixelours to \mpixel[].
Comparing the two directions, \mpixel[8L] and \mpixel[8S] hurt the initial languages to a similar extent:
these two variants perform on par on four of the five benchmarks.
Adding languages in a shared script therefore does not cost less than adding new scripts.
\mpixel[large] not only recovers these losses but achieves the best result on every benchmark, surpassing \mpixel[4S] and \pixelours on all five benchmarks.
Increased capacity therefore mitigates the curse in pixel models, similar to subword tokenizer-based models.%

\section{Further Analysis}
\label{sec:analysis}

The experimental results show that improvements are concentrated on the newly added languages and scripts (RQ1--RQ2), and that adding more languages reduces performance on the four languages shared by every model variant (RQ4). 
However, the results do not explain \emph{where} these effects arise in the model. 
In this section, we report the results of probing the hidden representations for their cross-lingual alignment and linguistic information.

\subsection{ABX-style Discrimination}
\label{subsec:abx}

\begin{figure}
    \centering
    \includegraphics[width=0.5\textwidth]{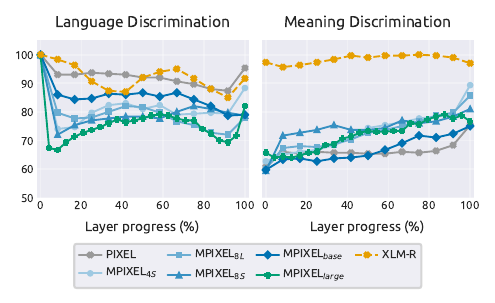}
    \caption{ABX probing performances of \mpixel[] variants on the four shared pretraining languages. }
    \label{fig:abx_all_models_4_languages}
\end{figure}

We use the ABX discrimination tasks proposed by
\citet{de-seyssel-etal-2025-discriminating} to measure the language and meaning representations of the models.
Given a triplet of
representations $(A, B, X)$, we measure whether $X$ is closer to $A$ or $B$ according to their cosine similarity.\footnote{Following \citet{de-seyssel-etal-2025-discriminating}, we use cosine similarity over mean-pooled sentence embeddings. For MD, we apply mean centering by subtracting the mean embedding of all sentences in the same language from each sentence representation, %
following \citet{libovicky-etal-2020-language}.} %
In the Meaning Discrimination (MD) task, we measure semantic similarity across languages while controlling for language identity, making it a proxy for cross-lingual semantic alignment. In the Language Discrimination (LD) task, we measure alignment between languages while controlling for meaning, measuring whether language-specific information is encoded in the representations. 
To cover all languages supported by our models, we use the multi-way parallel BOUQuET dataset \citep{andrews-etal-2025-bouquet}. 
Figure~\ref{fig:abx_all_models_4_languages} presents the results for ABX probes on the four shared languages in each \mpixel model.

\begin{figure*}[!t]
    \centering
    \includegraphics[width=\textwidth]{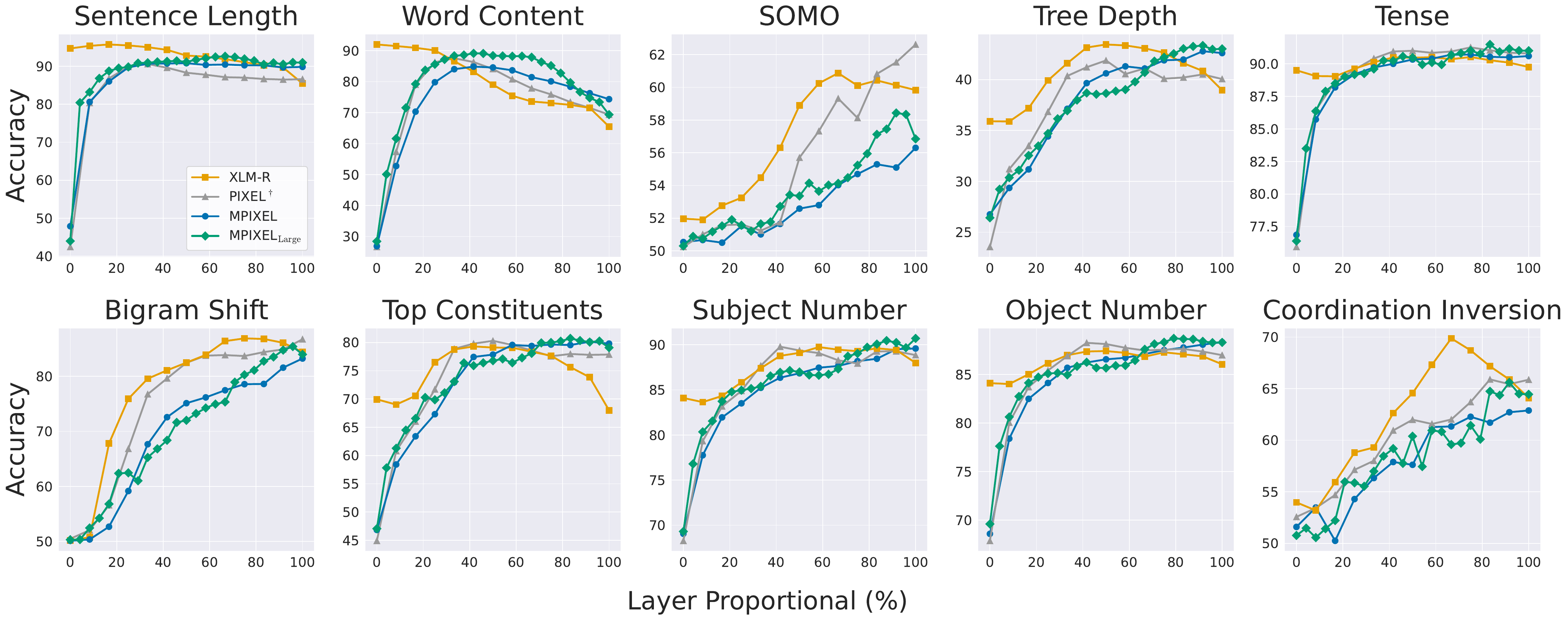}
    \caption{Layer-wise probing accuracy on sentence-level linguistic tasks. The accuracy is plotted against the proportional encoder layer depth for both base and large model variants.}
    \label{fig:sentence-level-probing}
\end{figure*}

\paragraph{Semantic alignment emerges with depth.} 
Across all base models, MD increases with depth and peaks in the final layer (Figure~\ref{fig:abx_all_models_4_languages}, right): representations become more semantically aligned across languages in higher layers, suggesting that the models gradually abstract away from visual, surface-level features toward more language-independent semantic representations. LD shows the complementary trend (Figure~\ref{fig:abx_all_models_4_languages}, left): all models start near-perfect in the embedding layer, which is expected given the different scripts and the resulting strong visual cues, but then drop sharply as the models move away from script-level features, before partially recovering in the middle layers. 

\paragraph{Adding languages and scripts weakens alignment.} Final-layer MD is ordered by the number of languages and scripts, \mpixel[4s]~$>$~\mpixel[8l]~$>$~\mpixel[8s]~$>$~\mpixel. Adding more languages and scripts therefore weakens cross-lingual semantic alignment, mirroring in the representations the curse of multilinguality we observe downstream \citep{conneau-etal-2020-unsupervised}. 

\paragraph{Effect of model capacity.} Comparing \mpixel and \mpixel[large](Figure~\ref{fig:abx_all_models_4_languages}), the larger model's MD scores are higher in the intermediate layers, showing that increased capacity improves cross-lingual alignment there. Over the final layers, however, its MD declines, so that \mpixel matches and even overtakes it in the last layer. This decline is driven specifically by same-script language pairs: their MD in \mpixel[large] behaves non-monotonically, rather than climbing steadily as it does in the smaller models. This pattern resembles what has been reported for subword models such as XLM-R (see \citet{de-seyssel-etal-2025-discriminating} and the XLM-R result in Figure~\ref{fig:abx_all_models_4_languages})\footnote{The MD results of XLM-R and \mpixel[large] are more in line with classical interpretations from previous work \citep{wu2025the, wendler-etal-2024-llamas} of middle layers as a \say{semantic hub}, where languages are more aligned.}, suggesting that the non-monotonicity stems from model capacity rather than from the pixel-based input. 

\paragraph{Surface similarity drives generalization.} Cross-lingual alignment also extends to languages held out of pretraining. Evaluating \mpixel and \mpixel[large] on four unseen languages, each sharing its script (and usually its family) with a seen language, the related seen--unseen pairs show the highest MD and the lowest LD at every depth, indicating that the model represents an unseen language similarly to its seen, same-script relative, indicating that alignment is a result of surface similarity rather than exposure (see Appendix~\ref{sec:appendix_abx} for details)\footnote{Relatedly, script information remains recoverable in all layers of \mpixel models, suggesting that they rely more on surface information even in deeper layers where tokenization-based models focus on semantics (see Appendix~\ref{sec:appendix_script_probes}).}.

\subsection{Sentence-level Probing}
\label{subsec:sentence_level_probing}

\citet{tatariya-etal-2024-pixology} established that pixel-based models perform visual processing in early layers before linguistic features emerge. 
We ask whether scaling model capacity alters this transition. 
Figure~\ref{fig:sentence-level-probing} shows the layer-wise accuracy of a trained mean-pooled probe for our \mpixel on the SentEval probing tasks~\citep{conneau-etal-2018-cram,conneau-kiela-2018-senteval}.
\paragraph{More layers are used for visual processing.}
We find that both the 12 and 24-layer models dedicate a proportionately similar number of layers to visual abstractions. 
On the surface tasks (SentLen and WC), 12-layer models plateau
between layers 3 and 4, at 25\% of the network depth, whereas
\mpixel[large] continues to improve until layer 9.

\paragraph{Syntax and surface semantics improve.}
This extended visual processing phase improves representations across several tasks. In the deeper layers of Tree Depth, \mpixel[large] outperforms both base models and narrows the gap with \xlmr. Furthermore, \mpixel[large] achieves the highest peak accuracy on all surface semantic tasks.
However, on complex semantics (SOMO), both \mpixel models fall behind the monolingual \pixelours, as SentEval is English-only.
Thus, scaling depth improves syntactic and surface semantic representations, but closing the semantic gap remains an open challenge.

\subsection{Word-level Probing}
\label{subsec:word_level_probing}
\begin{figure*}[!t]
    \centering
    \includegraphics[width=\textwidth]{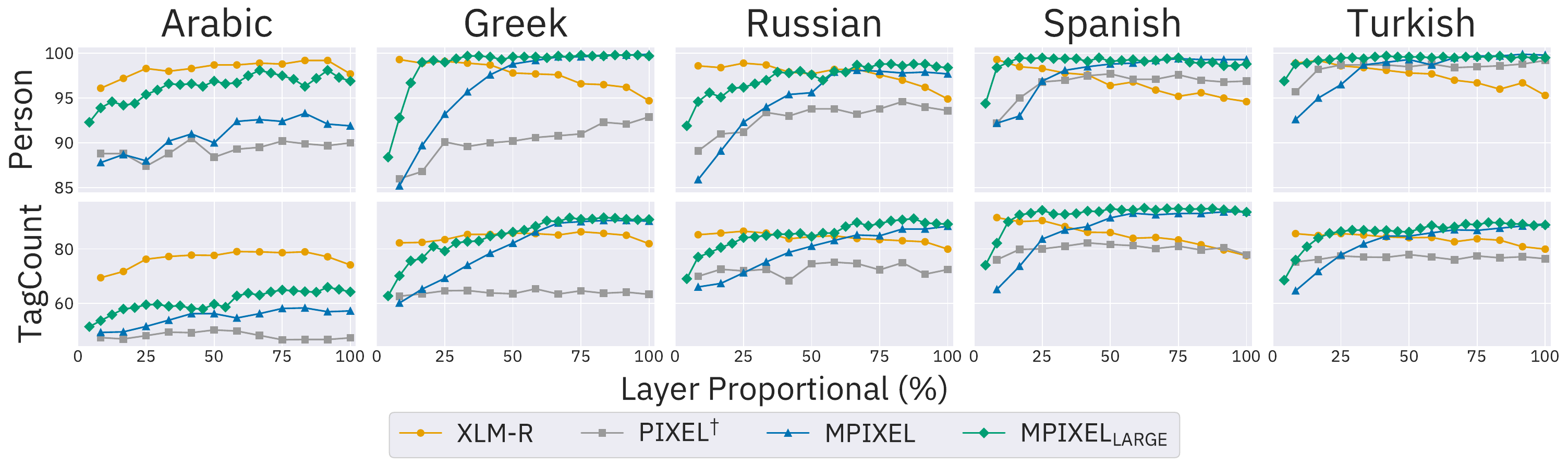}
    \caption{Layer-wise probing accuracy on \textsc{linspector} word-level tasks. Rows show tasks, columns show languages; each subplot plots accuracy against encoder layer.}
    \label{fig:linspector-main}
\end{figure*}

We probe the hidden representations of the models with \textsc{linspector}, a multilingual word-level probing benchmark~\citep{sahin-etal-2020-linspector}.
We train a linear classifier on the mean-pooled representations of each encoder layer and report accuracy on the test set.
We use two tasks:
Person requires predicting grammatical person, and TagCount requires predicting the number of morphological tags associated with a word.
TagCount is more difficult, as it requires capturing the amount of morphological information encoded in a word.
We evaluate on Arabic, Greek, Turkish, and Spanish, which are pretraining languages of \mpixel, and on Russian, which is typologically related to  Ukrainian.

The pixel models start low and improve with depth, while \xlmr reaches high accuracy already in the early layers (Figure~\ref{fig:linspector-main}).
This matches our ABX results and the findings of previous work \citep{tatariya-etal-2024-pixology,kesen-etal-2025-multilingual}: early pixel layers perform visual processing, and linguistic information emerges in deeper layers.
The difference with \xlmr varies by language: the \mpixel[] variants perform similar to \xlmr on Spanish and Turkish in deeper layers, but perform worse on Arabic on both tasks.
This may be due to the word-level rendering of Arabic, which we do not test.
\mpixel[large] closes the gap with \xlmr in the deeper layers, consistent with the extended visual phase seen in the sentence-level probes.

\section{Related Work}
\label{sec:related-work}
Visually rendered text processing was first applied to machine translation \citep{salesky-etal-2021-robust,salesky-etal-2023-multilingual}, replacing sub-word tokenization with images of the source language text.
\citet{rust2023language} introduced \textsc{pixel}, a language model built upon masked autoencoder architecture (ViT-MAE) and pretrained on visually rendered English text,
which outperformed its tokenizer-based equivalent \textsc{bert} \citep{devlin-etal-2019-bert} on languages with novel writing systems, together with increased robustness against orthographic perturbations.
\citet{lotz-etal-2023-text} demonstrated that structured rendering reduces patch input diversity, allowing a 25M parameter model to perform on par with the initial 86M parameter model; this approach was later applied to scanned historical documents \citep{borenstein-etal-2023-phd}.
Finally, there are hybrid approaches that  combine subword tokens and visually rendered text \citep{chai-etal-2024-autoregressive,gao2024improvinglanguageunderstandingscreenshots,lotz2025overcoming}, which finally enable language generation.

~\citet{kesen-etal-2025-multilingual} and~\citet{hu2026mixarscalingautoregressivepixelbased} are most relevant to our work on the  multilingual scaling of pixel language models.
\citet{kesen-etal-2025-multilingual} introduced \pixelmfour pretrained on four languages where each uses a distinct writing system, which demonstrated strong cross-lingual transfer learning performance to unseen scripts.
Subsequently, \citet{hu2026mixarscalingautoregressivepixelbased} extended patch-level autoregressive visual text generation model \textsc{pixar} \citep{tai-etal-2024-pixar} to eight languages.
In contrast to these works, we perform a controlled study of scaling pixel language models across languages, scripts, and model size under matched per-language compute.

Besides pixel language models, character- and byte-level models also offer vocabulary-free processing \citep{clark-etal-2022-canine,pagnoni-etal-2025-byte}.
Nonetheless, they have no access to the visual form of text, so orthographic similarities between characters and scripts remain unavailable to them.

\section{Conclusion}
\label{sec:conclusion}
We presented \mpixel{}, a family of pixel language models pretrained on up to sixteen languages and 307M parameters, scaled along three axes under fixed per-language compute: more languages in a seen script, more scripts, and more capacity.
We found that scaling acts locally: each axis mainly improves the languages and scripts it adds, while transfer to unseen scripts is established by the first multilingual step and does not improve with further scaling.
Pixel models exhibit the curse of multilinguality, and increased capacity mitigates it.
On low-resource languages typologically distant from the pretraining languages of all baselines, \mpixel{} outperforms subword and character-level baselines on many writing systems.
Our analyses showed that larger models reserve more layers for visual processing, and that cross-lingual alignment weakens with more languages but generalizes through shared scripts.
Future work will investigate explicit control over the capacity spent on visual processing, alternative rendering strategies for right-to-left scripts, and broader coverage of low-resource languages as scaling acts locally.

\section*{Limitations}
\label{sec:limitations}
Like most of the previous work, the pretrained \mpixel{} models lack language generation capability as they are encoder-only models.
Our scaling study covers few points per axis: four language counts (1, 4, 8, 16) and two model sizes (86M and 307M parameters), so our claims hold only within this range.
Due to compute constraints, our models are pretrained on far less data than \xlmr.
Moreover, each variant corresponds to a single pretraining run, so small differences between variants may fall within run-to-run pretraining variance, which we cannot quantify.
Our controlled setup fixes per-language compute rather than total compute, so variants with more languages see more total data, and results under a fixed total-compute budget may differ.
Each scaling axis is instantiated with a single set of languages, rooted in the four-language set of previous work for comparability, and our findings may be sensitive to these particular choices.
In particular, the first multilingual step adds new languages and new scripts at the same time, so we cannot separate their individual effects.
Finally, we design the \textsc{adaptive} rendering strategy based on input processing efficiency and correct visual rendering of right-to-left scripts, and we do not compare it empirically against existing rendering strategies such as \textsc{bigrams}.
Therefore, we do not know whether \textsc{adaptive} is the optimal rendering choice.

\section*{Ethical Considerations}
Like most pretrained language models, our models are pretrained on web-crawled corpora, which are already filtered and partly anonymized.
Nonetheless, these datasets might still contain some social biases, offensive content, and personal information, which our models may inherit as we do not apply any further filtering.
Our pretrained models lack generation capabilities, which limits their harmful use cases.
Finally, our models perform best on downstream tasks in their pretraining languages.
For other languages, we therefore recommend a careful evaluation before use.

\section*{Acknowledgements}
Ilker Kesen and Jenny Kunz were supported by the European Union’s Horizon 2020 research and innovation program under grant agreement No. 101135671 (TrustLLM). 
Desmond Elliott was supported by research grant (VIL53122) from VILLUM FONDEN and by the Carlsberg Foundation, grant CF25-1494.
The authors acknowledge the EuroHPC Joint Undertaking for awarding access to MareNostrum5, hosted at Barcelona Supercomputing Center (BSC), Spain, under proposals No. EHPC-DEV-2025D11-163 and EHPC-DEV-2025D11-030.

\bibliography{anthology-1,anthology-2,custom}

\appendix
\section{Appendix}
\subsection{Use of Frontier LLMs}
Large Language Models (LLMs) were used in the preparation of this work.
They were used to write and debug parts of the code, to produce the tables of experimental results and the plots, and to revise the text.
The code was reviewed and tested by the authors.
All written material was read and checked by all authors, who take full responsibility for the content of the work.

\subsection{Data Statistics}
\paragraph{Pretraining Data.}
Table~\ref{appendix:tab:pretraining-data-stats} reports the number of documents, words, and characters in the pretraining data sampled from FineWeb \citep{penedo2024fineweb} and FineWeb2 \citep{penedo2025fineweb} for each language.
The table also lists each language’s script, family, and branch.
Word counts are based on whitespace and are therefore omitted for Chinese and Thai.
Both FineWeb and FineWeb2 datasets are released under Open Data Commons Attribution License (ODC-By) v1.0.

\paragraph{Evaluation Benchmarks.}
SIB-200 \citep{adelani-etal-2024-sib} is a parallel text classification benchmark across 200 languages, where each language split includes 701, 99 and 204 examples for training, validation, and testing purposes, respectively.
SIB-200 is released under the CC BY-SA 4.0 license.
Table~\ref{tab:distant-language-list} lists the languages selected as typologically distant.
We used DistaLs~\citep{goot-etal-2025-distals} to calculate the average distance between each evaluation language and all baseline pretraining languages that use the same script.
We selected languages whose available distances all exceeded $T=0.50$, as well as languages written in scripts unseen during baseline pretraining.
Table~\ref{tab:appendix:udp-statistics} and Table~\ref{tab:ner-data-statistics} report the data statistics for the dependency parsing and NER tasks, respectively.
\textsc{linspector} is licensed under Apache 2.0, and each language split contains 7000, 2000, and 1000 examples for training, validation, and testing purposes, respectively.
GLUE benchmark statistics are reported in Table~\ref{tab:glue}.
For XNLI, we adopt the \texttt{translate-train-all} setting, in which the models are trained on 392,702 examples per language across all 15 languages ($\approx5.9M$ examples in total) and evaluated on 2,490 development and 5,010 test examples per language.

\begin{table}[t]
\centering
\resizebox{\columnwidth}{!}{
\begin{tabular}{lllr}
\toprule
Language & Script & Pivot & Distance \\
\midrule
\textsc{arb} & Latin & \textsc{hau} & 0.714 \\
\textsc{ayr} & Latin & \textsc{swh} & 0.725 \\
\textsc{bam} & Latin & \textsc{jav} & 0.609 \\
\textsc{cjk} & Latin & \textsc{swh} & 0.555 \\
\textsc{dyu} & Latin & \textsc{zsm} & 0.623 \\
\textsc{fij} & Latin & \textsc{sun} & 0.603 \\
\textsc{ibo} & Latin & \textsc{swh} & 0.531 \\
\textsc{nus} & Latin & \textsc{tam} & 0.732 \\
\textsc{quy} & Latin & \textsc{sun} & 0.691 \\
\textsc{taq} & Latin & \textsc{hau} & 0.610 \\
\textsc{umb} & Latin & \textsc{swh} & 0.516 \\
\textsc{wol} & Latin & \textsc{swh} & 0.638 \\
\textsc{yor} & Latin & \textsc{swh} & 0.661 \\
\textsc{ace} & Arabic & \textsc{pes} & 0.770 \\
\textsc{bjn} & Arabic & \textsc{arb} & 1.000 \\
\textsc{ckb} & Arabic & \textsc{pbt} & 0.626 \\
\textsc{kas} & Arabic & \textsc{urd} & 0.525 \\
\textsc{mni} & Bengali & \textsc{ben} & 0.759 \\
\textsc{tgk} & Cyrllic & \textsc{mkd} & 0.687 \\
\textsc{shn} & Myanmar & \textsc{mya} & 0.781 \\
\textsc{nqo} & N'Ko & -- & -- \\
\textsc{sat} & Ol Chiki & -- & -- \\
\textsc{taq} & Tifinagh & -- & -- \\
\textsc{tzm} & Tifinagh & -- & -- \\
\textsc{bod} & Tibetan & -- & -- \\
\textsc{dzo} & Tibetan & -- & -- \\
\bottomrule
\end{tabular}}
\caption{Language--script pairs selected as typologically distant (DistaLs average of typology-related distances, $T=0.50$). The nearest same-script pretraining pivot and distance are shown.}
\label{tab:distant-language-list}
\end{table}

\begin{table*}[!ht]
\centering
\resizebox{\textwidth}{!}{%
\begin{tabular}{@{}lllllrrr@{}}
\toprule
ISO & Language & Script & Language Family & Branch & \# Docs (M) & \# Words (B) & \# Chars (B) \\
\midrule
\textsc{ben} & Bengali & Bengali & Indo-European & Indo-Aryan & 6.9 & 2.5 & 13.9 \\
\textsc{dan} & Danish & Latin & Indo-European & Germanic & 16.7 & 2.3 & 12.2 \\
\textsc{eng} & English & Latin & Indo-European & Germanic & 17.0 & 2.5 & 12.2 \\
\textsc{fin} & Finnish & Latin & Uralic & Finnic & 16.4 & 1.8 & 13.0 \\
\textsc{heb} & Hebrew & Hebrew & Afro-Asiatic & Semitic & 17.0 & 2.5 & 11.7 \\
\textsc{hin} & Hindi & Devanagari & Indo-European & Indo-Aryan & 18.6 & 3.7 & 15.0 \\
\textsc{kor} & Korean & Hangul & Koreanic & -- & 17.6 & 2.2 & 7.7 \\
\textsc{cmn} & Mandarin Chinese & Han & Sino-Tibetan & Sinitic & 17.2 & -- & 7.5 \\
\textsc{ell} & Modern Greek & Greek & Indo-European & Hellenic & 17.1 & 2.2 & 12.2 \\
\textsc{fas} & Persian & Arabic & Indo-European & Iranian & 17.6 & 3.7 & 15.0 \\
\textsc{spa} & Spanish & Latin & Indo-European & Romance & 16.9 & 2.5 & 12.5 \\
\textsc{arb} & Standard Arabic & Arabic & Afro-Asiatic & Semitic & 17.9 & 3.2 & 15.8 \\
\textsc{tha} & Thai & Thai & Kra-Dai & Tai & 17.2 & -- & 14.4 \\
\textsc{tur} & Turkish & Latin & Turkic & Oghuz & 17.0 & 1.9 & 12.6 \\
\textsc{ukr} & Ukrainian & Cyrillic & Indo-European & Slavic & 16.9 & 1.9 & 11.7 \\
\textsc{vie} & Vietnamese & Latin & Austroasiatic & Vietic & 16.2 & 3.1 & 11.1 \\
\midrule
Total & & & & & 264M & $>$36B& 211B \\
\bottomrule
\end{tabular}%
}
\caption{Per-language statistics of the pretraining corpus: script, language family and branch, and corpus size in documents (M, millions), whitespace-delimited words (B, billions) and characters (B). Word counts are omitted for languages without whitespace word boundaries (Chinese, Thai).}
\label{appendix:tab:pretraining-data-stats}
\end{table*}

\begin{table*}[ht]
\centering
\resizebox{0.6\textwidth}{!}{%
\begin{tabular}{@{}llrl@{}}
\toprule
Language & Treebank           & \#Sentences                  & License         \\ \midrule
\textsc{eng}     & English-EWT        & 16621                        & CC BY-SA 4.0    \\
\textsc{ara}     & Arabic-PADT        & 7664                         & CC BY-NC-SA 3.0 \\
\textsc{hye} & Armenian-BSUT & 2523 & CC BY-SA 4.0 \\
\textsc{bul}     & Bulgarian-BTB      & 11138 & CC BY-NC-SA 3.0 \\
\textsc{cop}     & Coptic-Scriptorium & 2011                         & CC BY 4.0       \\
\textsc{dan} & Danish-DDT & 5512 & CC BY-SA 4.0 \\
\textsc{est} & Estonian-EWT & 7190 & CC BY-NC-SA 4.0 \\
\textsc{fao} & Faroese-FarPaHC & 1621 & CC BY-SA 4.0 \\
\textsc{fin} & Finnish-FTB & 18723 & CC BY 4.0 \\
\textsc{kat} & Georgian-GLC & 3164 & CC BY-SA 4.0 \\
\textsc{ell} & Greek-GDT & 2521 & CC BY-NC-SA 3.0 \\
\textsc{heb} & Hebrew-HTB & 6143 & CC BY-NC-SA 4.0 \\
\textsc{hin}     & Hindi-HDTB         & 16647                        & CC BY-NC-SA 4.0 \\
\textsc{ind} & Indonesian-GSD & 5598 & CC BY-SA 4.0 \\
\textsc{jpn}     & Japanese-GSD       & 8100                         & CC BY-SA 4.0    \\
\textsc{kor}     & Korean-GSD         & 6339                         & CC BY-SA 4.0    \\
\textsc{mlt} & Maltese-MUDT & 2074 & CC BY-SA 4.0 \\
\textsc{fas} & Persian-Seraji & 5997 &  CC BY-SA 4.0 \\
\textsc{rus}     & Russian-GSD        & 5030                         & CC BY-SA 4.0    \\
\textsc{spa} & Spanish-GSD & 16013 & CC BY-SA 4.0 \\
\textsc{tam}     & Tamil-TTB          & 600                          & CC BY-NC-SA 3.0 \\
\textsc{tel}     & Telugu-MTG         & 5130                         & CC BY-SA 4.0    \\
\textsc{tha} & Thai-TUD & 3627 & CC BY-SA 4.0 \\
\textsc{tur} & Turkish-BOUN & 9761 & CC BY-SA 4.0 \\
\textsc{ukr}     & Ukrainian-IU       & 5030                         & CC BY-NC-SA 4.0 \\
\textsc{urd}     & Urdu-UDTB          & 5130                         & CC BY-NC-SA 4.0 \\
\textsc{uig} & Uyghur-UDT & 3456 & CC BY-SA 4.0 \\
\textsc{vie}     & Vietnamese-VTB     & 3000                         & CC BY-SA 4.0    \\
\textsc{wol} & Wolof-WTB & 2107 & CC BY-SA 4.0 \\
\textsc{zho}     & Chinese-GSD        & 4997                         & CC BY-SA 4.0    \\
\bottomrule
\end{tabular}%
}
\caption{Total number of sentences of Universal Dependencies v2.18 \citep{udp2.18,nivre-etal-2020-universal} treebanks used for dependency parsing task evaluations, including dataset licenses. Adapted from \citet{rust2023language,kesen-etal-2025-multilingual}.}
\label{tab:appendix:udp-statistics}
\end{table*}

\begin{table*}[!htbp]
 \centering
 \resizebox{\textwidth}{!}{%
  \begin{tabular}{lrrrll}
    \toprule
    Language & Train & Validation & Test & Source & License \\
    \midrule
    \textsc{ben} & 961679 & 4859 & 607 & Naamapadam & CC0 1.0 \\
    \textsc{dan} & 4383 & 564 & 565 & Universal NER (UNER\_Danish-DDT) & CC BY-SA 4.0 \\
    \textsc{ell} & 1662 & 403 & 456 & Universal NER (UNER\_Greek\_GDT) & CC-BY-SA-4.0 \\
    \textsc{eng} & 12543 & 2001 & 2077 & Universal NER (UNER\_English-EWT) & CC BY-SA 4.0 \\
    \textsc{heb} & 5241 & 484 & 491 & Universal NER (UNER\_Hebrew-HTB) & CC BY-SA 4.0 \\
    \textsc{hin} & 985787 & 13460 & 867 & Naamapadam & CC0 1.0 \\
    \textsc{kor} & 21008 & 5000 & -- & KLUE & CC BY-SA 4.0 \\
    \textsc{srp} & 3328 & 536 & 520 & Universal NER (UNER\_Serbian-SET) & CC BY-SA 4.0 \\
    \textsc{tam} & 497882 & 2795 & 758 & Naamapadam & CC0 1.0 \\
    \textsc{tel} & 507741 & 2700 & 847 & Naamapadam & CC0 1.0 \\
    \textsc{ukr} & 10980 & 1206 & 5593 & NER-UK 2.0 & CC BY-NC-SA 4.0 \\
    \textsc{zho} & 3997 & 500 & 500 & Universal NER & CC BY-SA 4.0 \\
    \bottomrule
  \end{tabular}}
\caption{Data distribution and license information for the NER datasets used within this work. See Universal NER \citep{mayhew-etal-2024-universal,blevins-etal-2026-universal}, Naamapadam \citep{mhaske-etal-2023-naamapadam}, NER-UK \citep{chaplynskyi-romanyshyn-2024-introducing} and KLUE \citep{park2021klue} for more detailed statistics.}
\label{tab:ner-data-statistics}
\end{table*}

\subsection{Implementation Details}

Tables~\ref{tab:pretraining-hyperparams-shared} and~\ref{tab:pretraining-hyperparams-different} contain the hyperparameters used to pretrain the \mpixel models.
Table~\ref{tab:evaluation-hyperparams} contains the hyperparameters used to finetune and evaluate the models on the downstream tasks.
The learning rates depend on the task and the model size. For SIB-200, UDP and NER, we use $\{$1e$-$5, 3e$-$5, 5e$-$5$\}$ for both base and large models. For \mpixel[large] on UDP, we also use 5e$-$6.
For XNLI, we use $\{$8e$-$6, 1e$-$5, 2e$-$5, 3e$-$5, 5e$-$5$\}$ for base models, and $\{$5e$-$6, 8e$-$6, 1e$-$5, 2e$-$5, 3e$-$5$\}$ for \mpixel[large].
For GLUE, we use 1e$-$5 for WNLI~\citep{levesque2012winograd}, 2e$-$5 for STS-B~\citep{cer-etal-2017-semeval}, and 3e$-$5 for MRPC~\citep{dolan-brockett-2005-automatically}, QNLI~\citep{wang-etal-2018-glue}, QQP, RTE and SST-2~\citep{socher-etal-2013-recursive}.
For CoLA~\citep{warstadt-etal-2019-neural}, we use $\{$1e$-$5, 2e$-$5$\}$, and for MNLI~\citep{williams-etal-2018-broad}, we use $\{$1e$-$5, 2e$-$5, 3e$-$5$\}$.
For large models, we also use 5e$-$6 for WNLI, and 1e$-$5 for STS-B, MRPC, QNLI, QQP, RTE and SST-2.
Following \citet{rust2023language,kesen-etal-2025-multilingual}, for each learning rate, we run several experiments with different random seeds. From all these runs, we take the $3$ best ones and report their average score.

We adopt Flash Attention 2 \citep{dao2023flashattention2} to reduce memory usage and accelerate attention computation, and use \texttt{torch.compile} to further speed up forward and backward passes. 
With these optimizations, pretraining takes approximately 1350 GPU hours for \mpixel and 2500 GPU hours for \mpixel[large] on NVIDIA H100s.

\begin{table*}[!htbp]
\centering
\resizebox{0.8\textwidth}{!}{%
\begin{tabular}{@{}ll@{}}
\toprule
Parameter            & Value \\ \midrule
Image size                    & (16, 8192, 3)  \\
Patch size $P$                   & 16             \\
Layer norm $\varepsilon$ \citep{ba2016layernormalization}       & \num{1e-12}    \\
Span masking ratio $R$                & 0.25           \\
Span masking max length $S$ & 6 \\
Span masking cumulative weights $W$ & $\{0.2, 0.4, 0.6, 0.8, 0.9, 1\}$ \\
Span masking spacing & Dynamic \\
Dropout probability           & 0.1            \\ 
Hidden activation & GeLU \citep{DBLP:journals/corr/HendrycksG16} \\
Optimizer              & AdamW \citep{loshchilov2018decoupled, kingma2017adammethodstochasticoptimization}                \\
Adam $\beta_1$     & (0.9) \\
Adam $\beta_2$     & $\{0.999,  0.95\}$ \\
Adam $\varepsilon$       & \num{1e-8}     \\
Weight decay               & 0.05           \\
Learning rate schedule & Cosine Decay \citep{DBLP:conf/iclr/LoshchilovH17}                         \\
Minimum learning rate      & \num{1e-5}     \\
Learning rate warmup ratio & 0.05           \\
Training steps             & 1M             \\
\bottomrule
\end{tabular}
}
\caption{Pretraining hyperparameters shared by all \mpixel models and \pixelours.}
\label{tab:pretraining-hyperparams-shared}
\end{table*}

\begin{table*}[!htbp]
\centering
\resizebox{0.8\textwidth}{!}{%
\begin{tabular}{lll}
\hline
Parameter                     & Base Models    & \mpixel[large] \\ \hline
Encoder hidden size $D_{\text{enc}}$ & 768                             & 1024           \\
Encoder intermediate size            & 3072                            & 4096           \\
Encoder num attention heads          & 12                              & 16             \\
Encoder num layers $L$               & 12                              & 24             \\
Decoder hidden size $D_{\text{dec}}$ & 512                             & 512            \\
Decoder intermediate size            & 2048                            & 2048           \\
Decoder num attention heads          & 16                              & 16             \\
Decoder num layers $K$               & 8                               & 8              \\
Learning rate                        & $\{1.5e^-4, ~2.12e^-4,~ 3e^-4\}$ & 3e-4           \\
Batch size                           & $\{256,~512,~ 1024\}$            & 1024           \\ \hline
\end{tabular}}
\caption{Pretraining hyperparameters that differ between  base-scale models with 86M parameters and the large \mpixel[] model with 307M parameters.}
\label{tab:pretraining-hyperparams-different}
\end{table*}

\begin{table*}[t]
\centering
\resizebox{0.7\textwidth}{!}{%
\begin{tabular}{@{}lccccc@{}}
\toprule
\textbf{Parameter} & \textbf{SIB-200} & \textbf{UDP} & \textbf{NER} & \textbf{XNLI} & \textbf{GLUE} \\
\midrule
Classification head pooling & Mean & --- & --- & CLS & Mean \\
Optimizer & \multicolumn{5}{c}{AdamW} \\
Adam $\beta$ & \multicolumn{5}{c}{0.9, 0.999} \\
Adam $\epsilon$ & \multicolumn{5}{c}{1e$-$8} \\
Weight decay & \multicolumn{5}{c}{0} \\
Learning rate schedule & \multicolumn{5}{c}{Linear decay} \\
Warmup steps & 100 & 100 & 100 & 1000 & 100 \\
Max sequence length & 256 & 256 & 256 & 256 & 256 \\
Batch size & 64 & 64 & 64 & 256 & $\{64, 256\}$ \\
Max steps & 15\,000 & 15\,000 & 15\,000 & 50\,000 & 400, 15\,000 \\
Eval strategy & epochs & steps & steps & steps & epochs, steps \\
Eval steps & --- & 100, 500 & 500 & 1000 & 10, 100, 250, 500 \\
Early stopping & \multicolumn{5}{c}{\checkmark} \\
Early stopping patience & 20 & 5, 20 & 5 & 5 & 5, 50 \\
Dropout probability & \multicolumn{5}{c}{0.1} \\
\bottomrule
\end{tabular}}
\caption{Hyperparameters used for fine-tuning and evaluating models on SIB-200, UDP, NER, XNLI and GLUE.}
\label{tab:evaluation-hyperparams}
\end{table*}

\subsection{Full Results}
\label{sec:appendix:results}

Table~\ref{tab:sib} shows the complete per-language results of the SIB-200 experiments.
Table~\ref{tab:ner} shows the complete per-language results of the NER experiments.
Table~\ref{tab:xnli-test} shows the complete per-language results of the XNLI experiments.
Table~\ref{tab:glue} shows the complete results of the GLUE experiments.
Table~\ref{tab:udp} shows the complete per-language results of the Universal Dependencies Parsing experiments.

\subsection{Full LINSPECTOR Results}
\label{sec:appendix:linspector}
\begin{figure*}[!htbp]
  \centering
  \includegraphics[width=\textwidth]{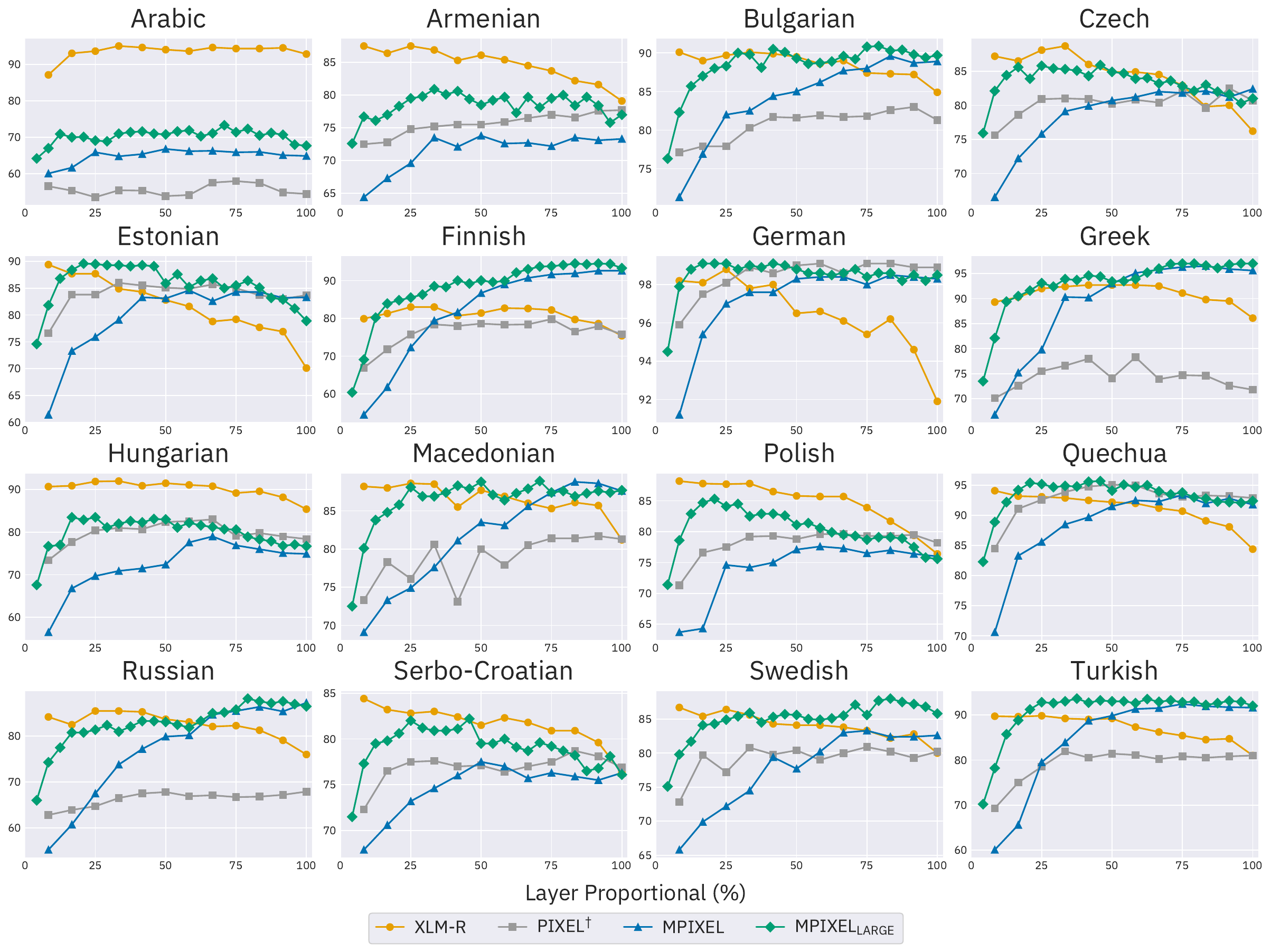}
  \caption{\textsc{linspector} probing results for the \textit{Case} task. Each panel shows linear probe accuracy per layer, with the x-axis giving the layer position relative to the model depth.}
  \label{fig:linspector-case}
\end{figure*}

\begin{figure*}[!htbp]
  \centering
  \includegraphics[width=\textwidth]{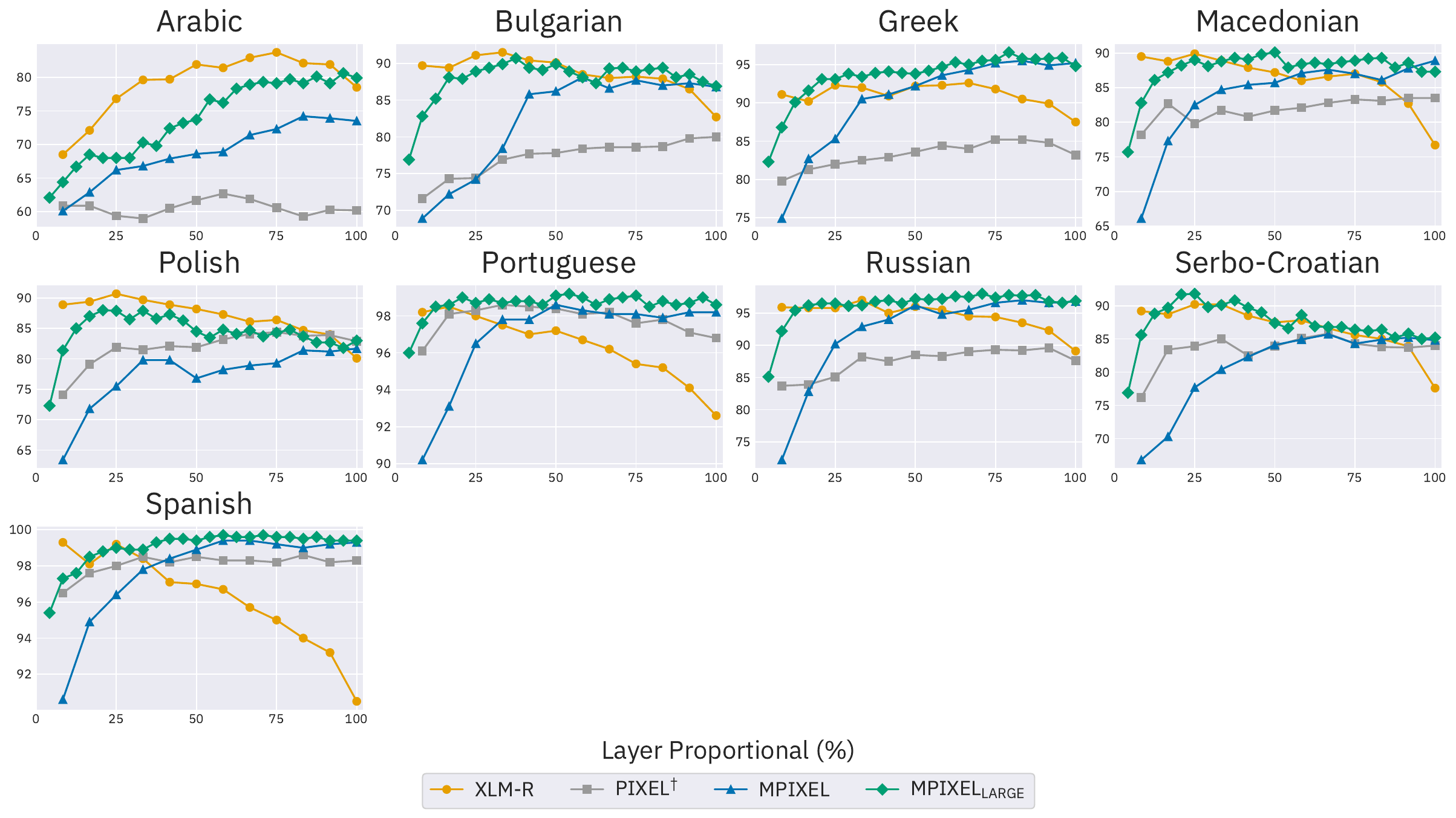}
  \caption{\textsc{linspector} probing results for the \textit{Gender} task. Each panel shows linear probe accuracy per layer, with the x-axis giving the layer position relative to the model depth.}
  \label{fig:linspector-gender}
\end{figure*}

\begin{figure*}[!htbp]
  \centering
  \includegraphics[width=0.9\textwidth]{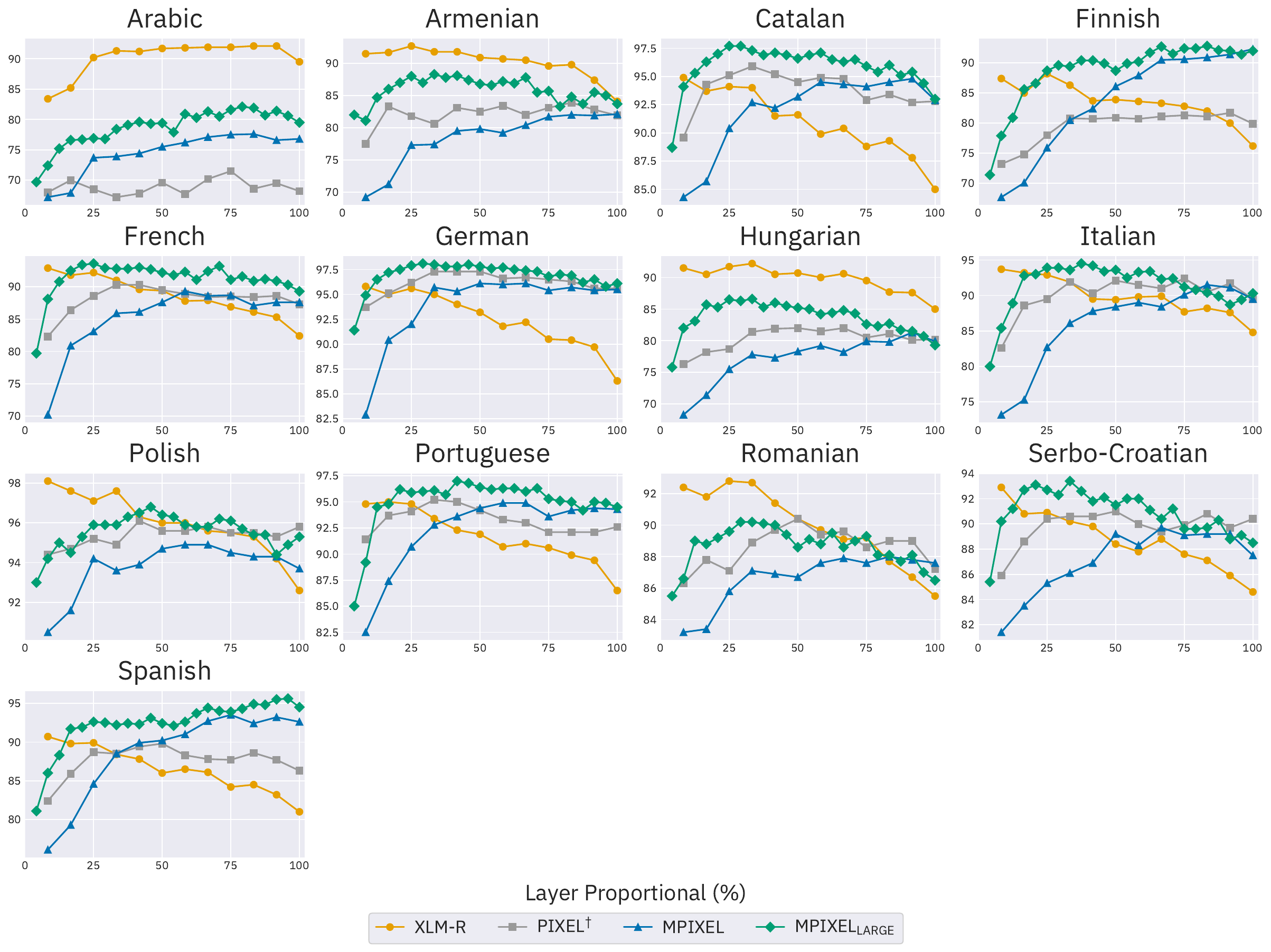}
  \caption{\textsc{linspector} probing results for the \textit{Mood} task. Each panel shows linear probe accuracy per layer, with the x-axis giving the layer position relative to the model depth.}
  \label{fig:linspector-mood}
\end{figure*}

\begin{figure*}[!htbp]
  \centering
  \includegraphics[width=0.9\textwidth]{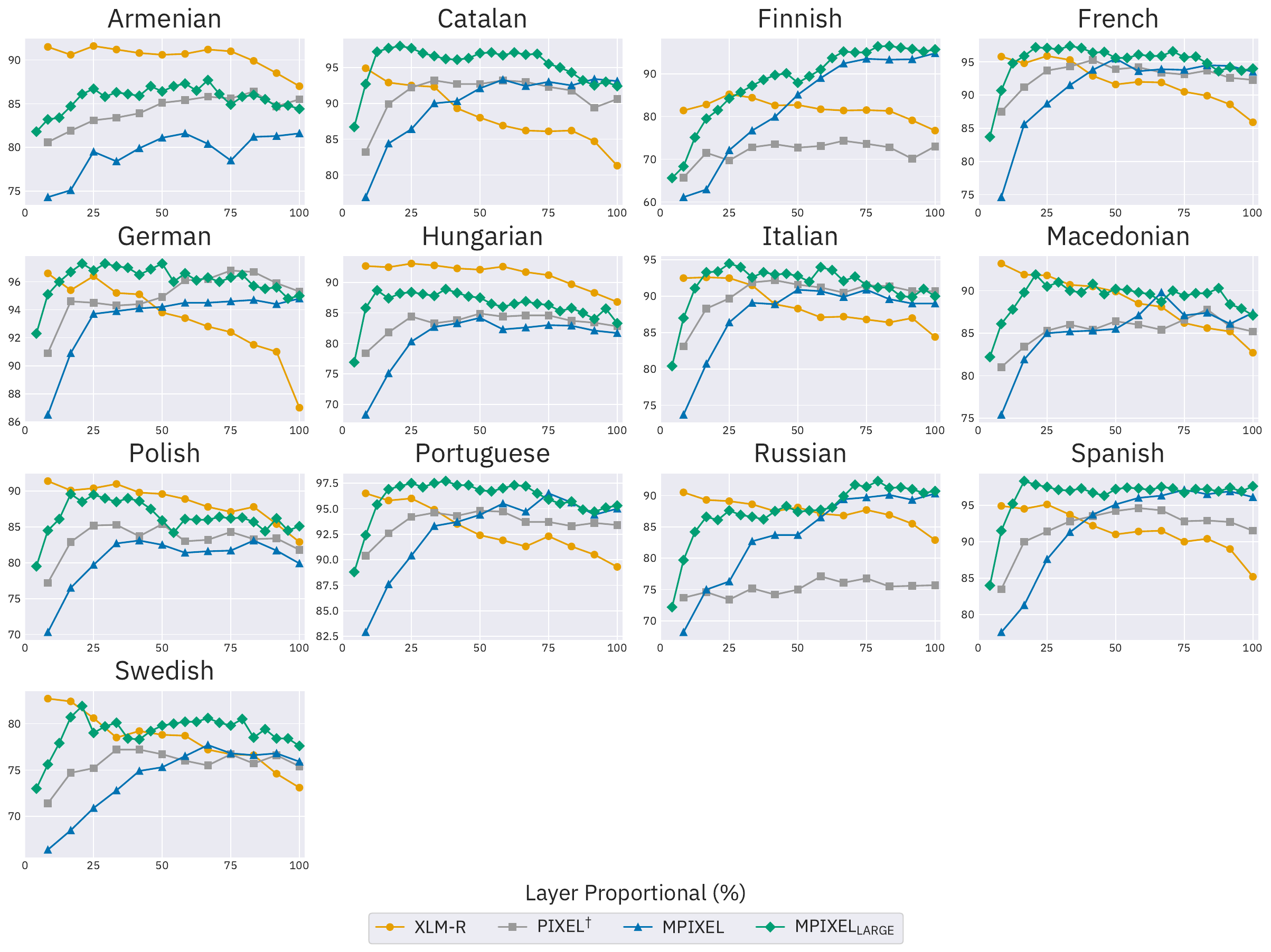}
  \caption{\textsc{linspector} probing results for the \textit{Number} task. Each panel shows linear probe accuracy per layer, with the x-axis giving the layer position relative to the model depth.}
  \label{fig:linspector-number}
\end{figure*}

\begin{figure*}[!htbp]
  \centering
  \includegraphics[width=\textwidth]{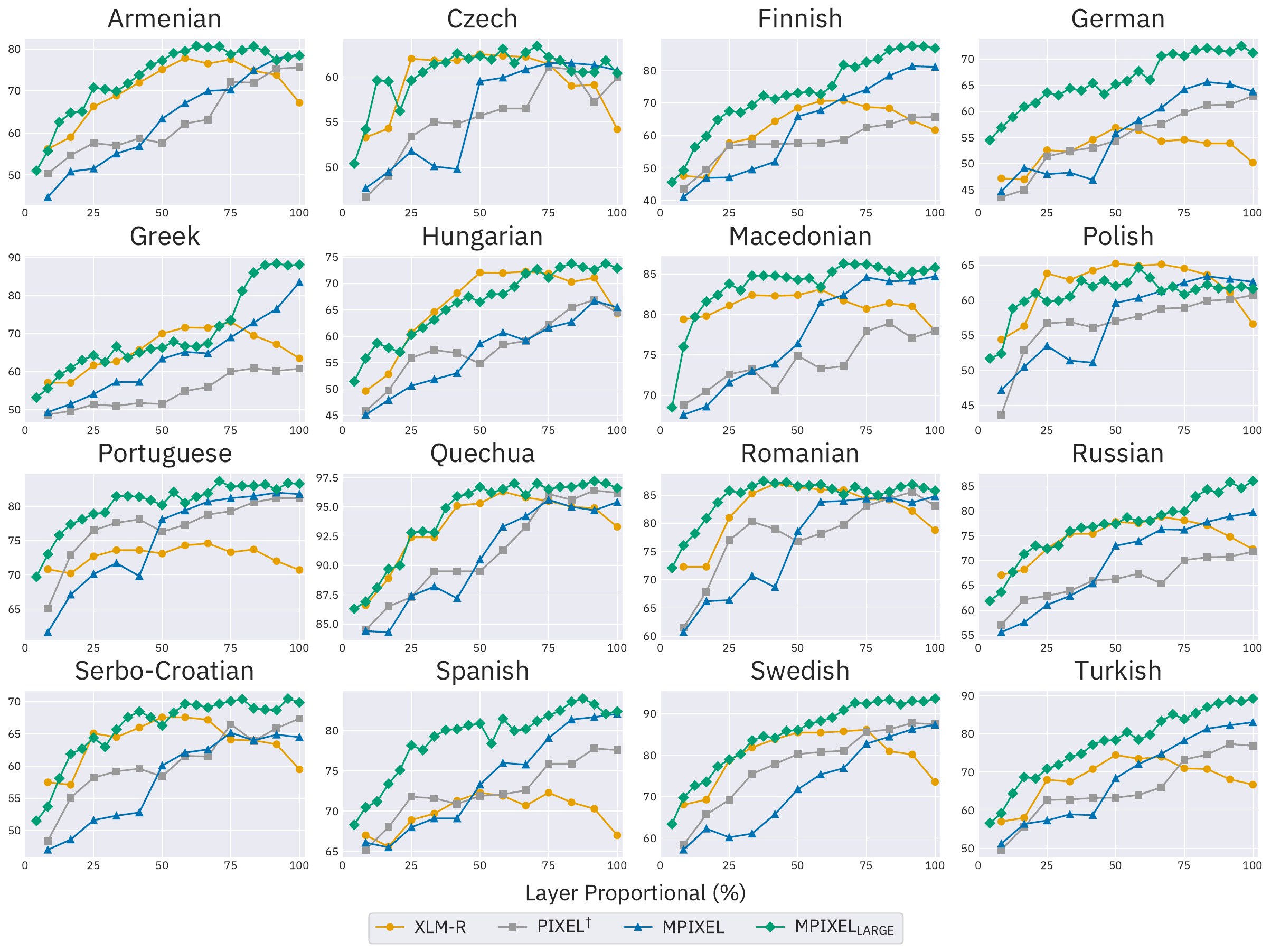}
  \caption{\textsc{linspector} probing results for the \textit{OddFeat} task. Each panel shows linear probe accuracy per layer, with the x-axis giving the layer position relative to the model depth.}
  \label{fig:linspector-oddfeat}
\end{figure*}

\begin{figure*}[!htbp]
  \centering
  \includegraphics[width=\textwidth]{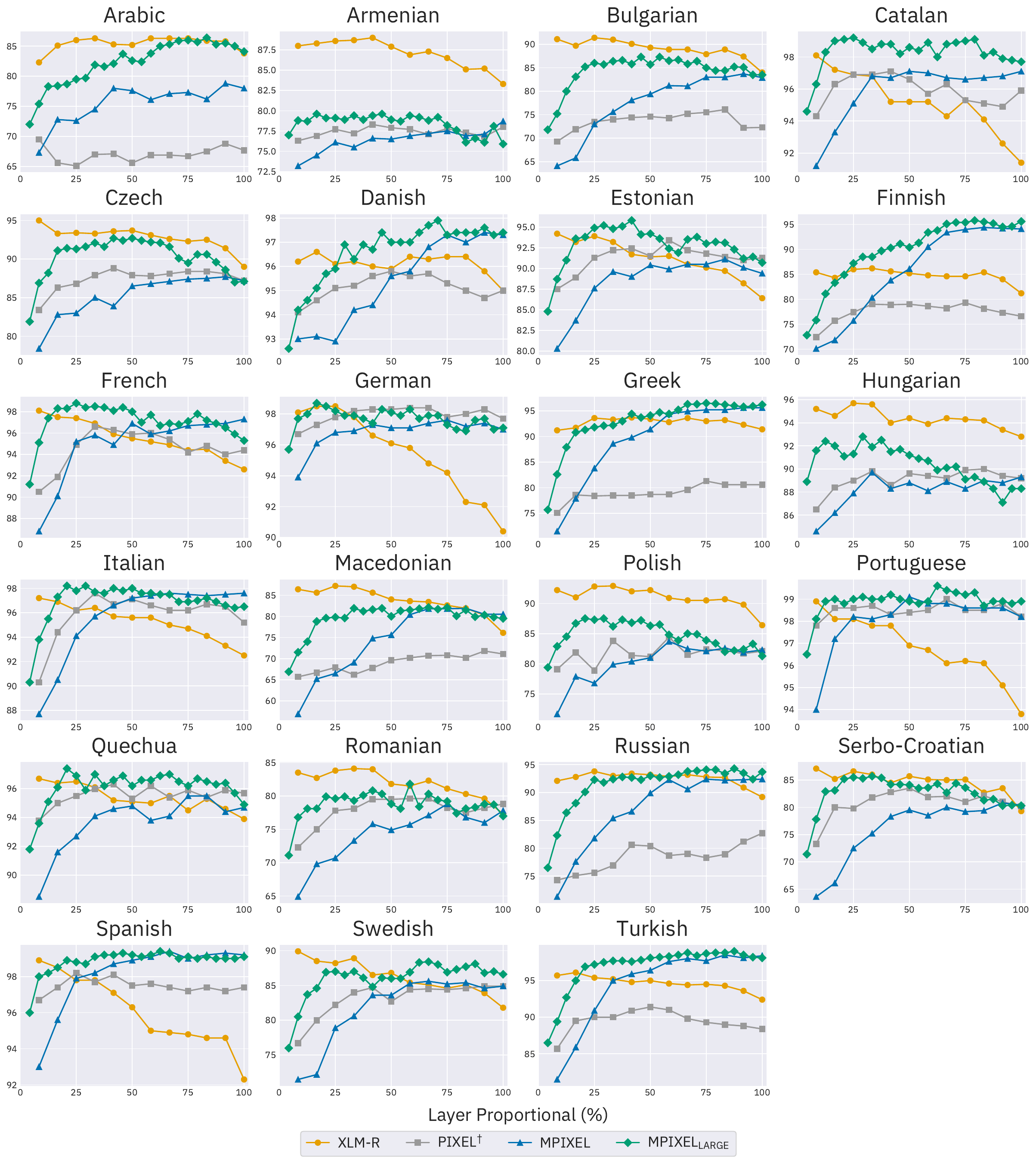}
  \caption{\textsc{linspector} probing results for the \textit{POS} task. Each panel shows linear probe accuracy per layer, with the x-axis giving the layer position relative to the model depth.}
  \label{fig:linspector-pos}
\end{figure*}

\begin{figure*}[!htbp]
  \centering
  \includegraphics[width=\textwidth]{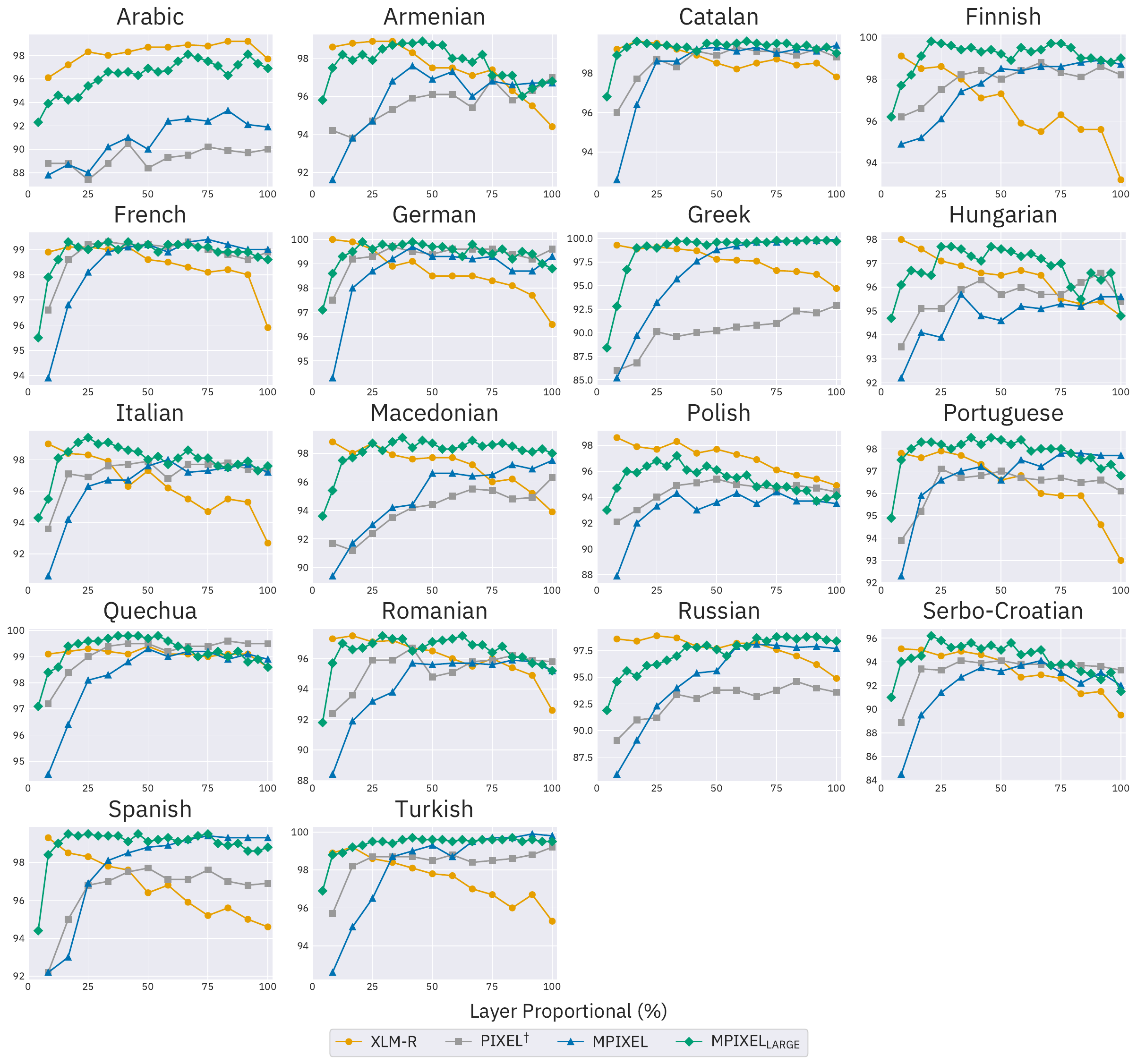}
  \caption{\textsc{linspector} probing results for the \textit{Person} task. Each panel shows linear probe accuracy per layer, with the x-axis giving the layer position relative to the model depth.}
  \label{fig:linspector-person}
\end{figure*}

\begin{figure*}[!htbp]
  \centering
  \includegraphics[width=0.75\textwidth]{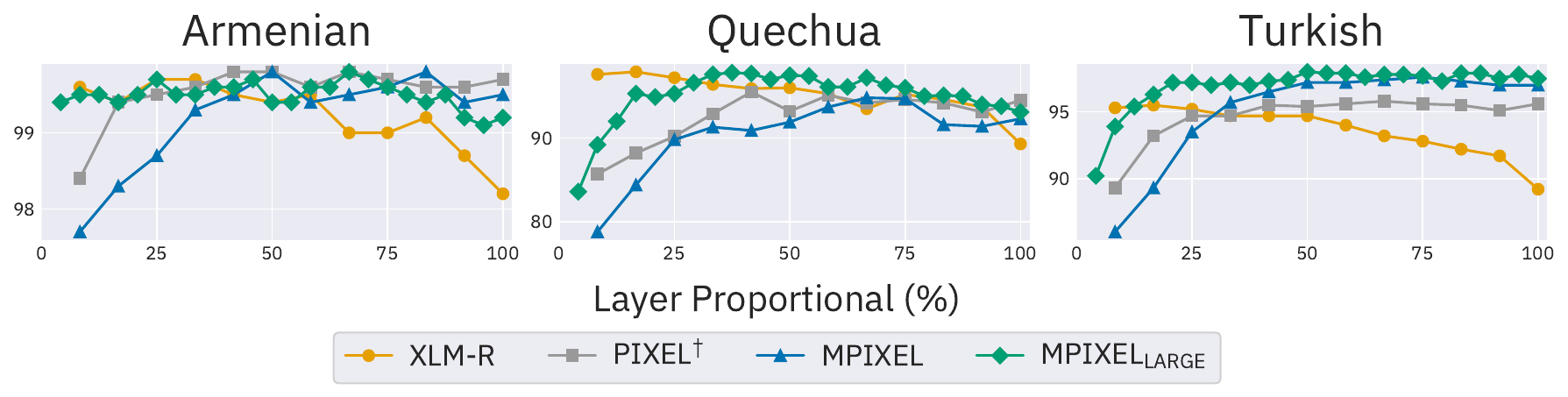}
  \caption{\textsc{linspector} probing results for the \textit{Possession} task. Each panel shows linear probe accuracy per layer, with the x-axis giving the layer position relative to the model depth.}
  \label{fig:linspector-possession}
\end{figure*}

\begin{figure*}[!htbp]
  \centering
  \includegraphics[width=\textwidth]{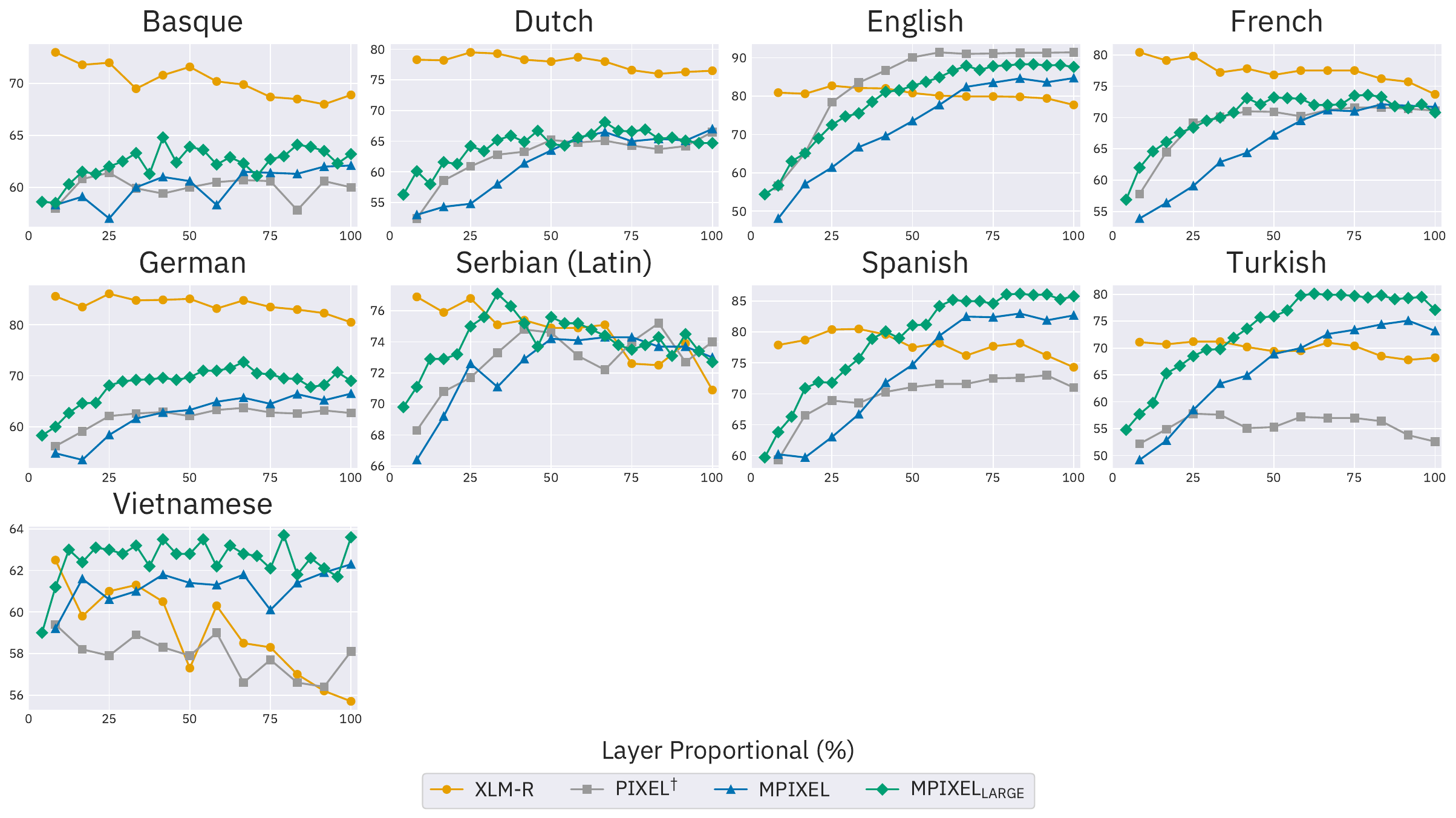}
  \caption{\textsc{linspector} probing results for the \textit{Pseudo} task. Each panel shows linear probe accuracy per layer, with the x-axis giving the layer position relative to the model depth.}
  \label{fig:linspector-pseudo}
\end{figure*}

\begin{figure*}[!htbp]
  \centering
  \includegraphics[width=\textwidth]{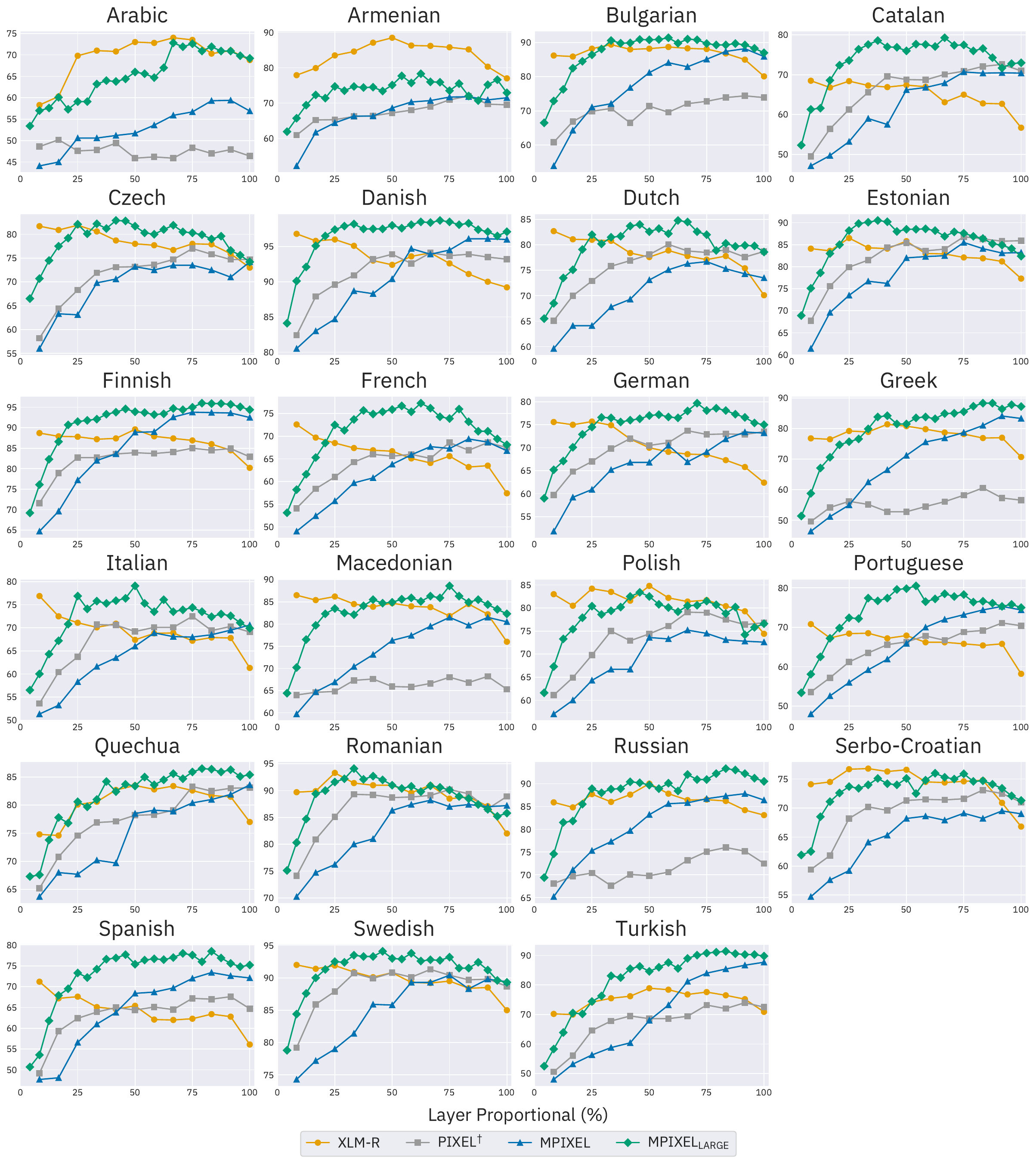}
  \caption{\textsc{linspector} probing results for the \textit{SameFeat} task. Each panel shows linear probe accuracy per layer, with the x-axis giving the layer position relative to the model depth.}
  \label{fig:linspector-samefeat}
\end{figure*}

\begin{figure*}[!htbp]
  \centering
  \includegraphics[width=\textwidth]{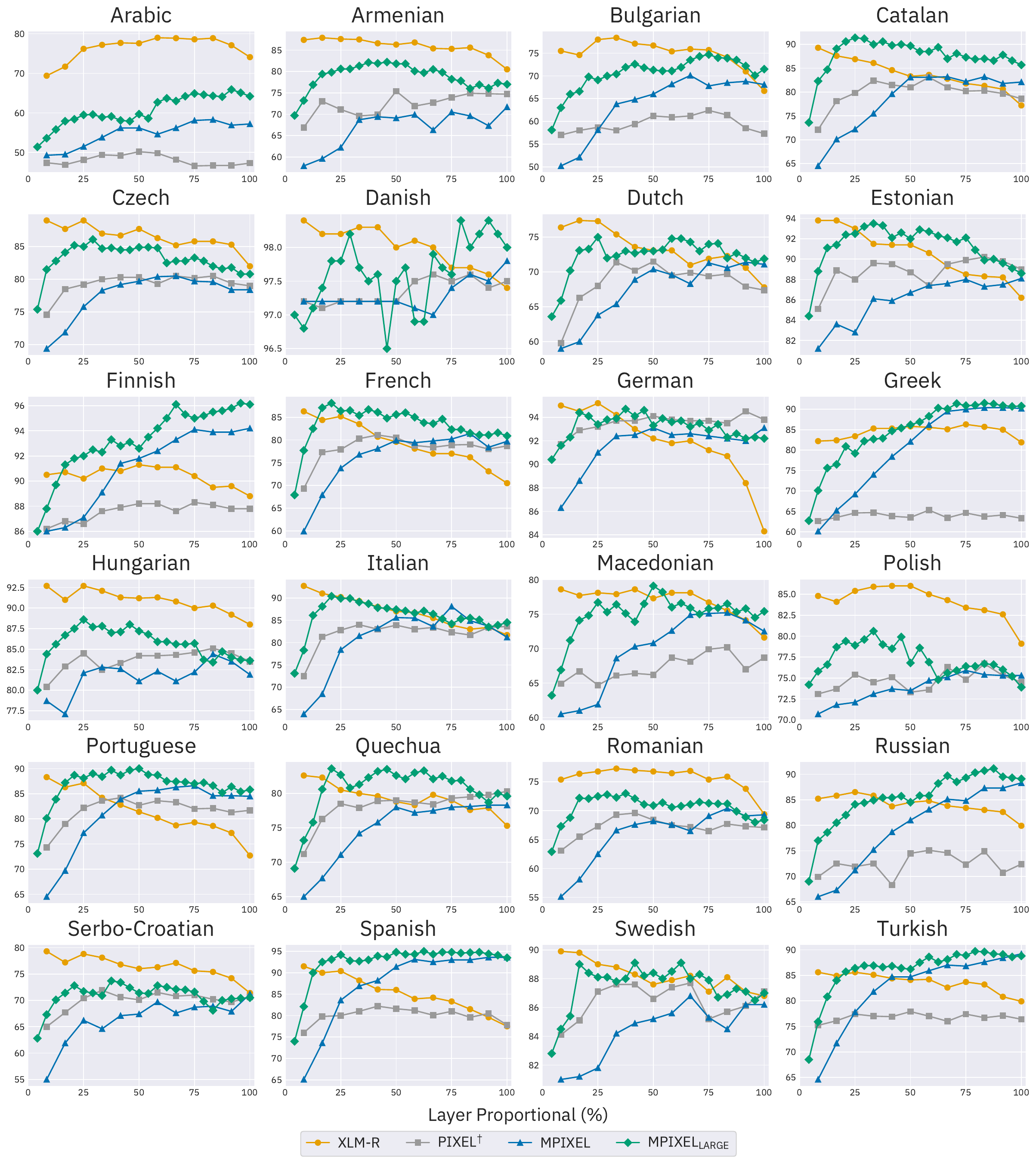}
  \caption{\textsc{linspector} probing results for the \textit{TagCount} task. Each panel shows linear probe accuracy per layer, with the x-axis giving the layer position relative to the model depth.}
  \label{fig:linspector-tagcount}
\end{figure*}

\begin{figure*}[!htbp]
  \centering
  \includegraphics[width=\textwidth]{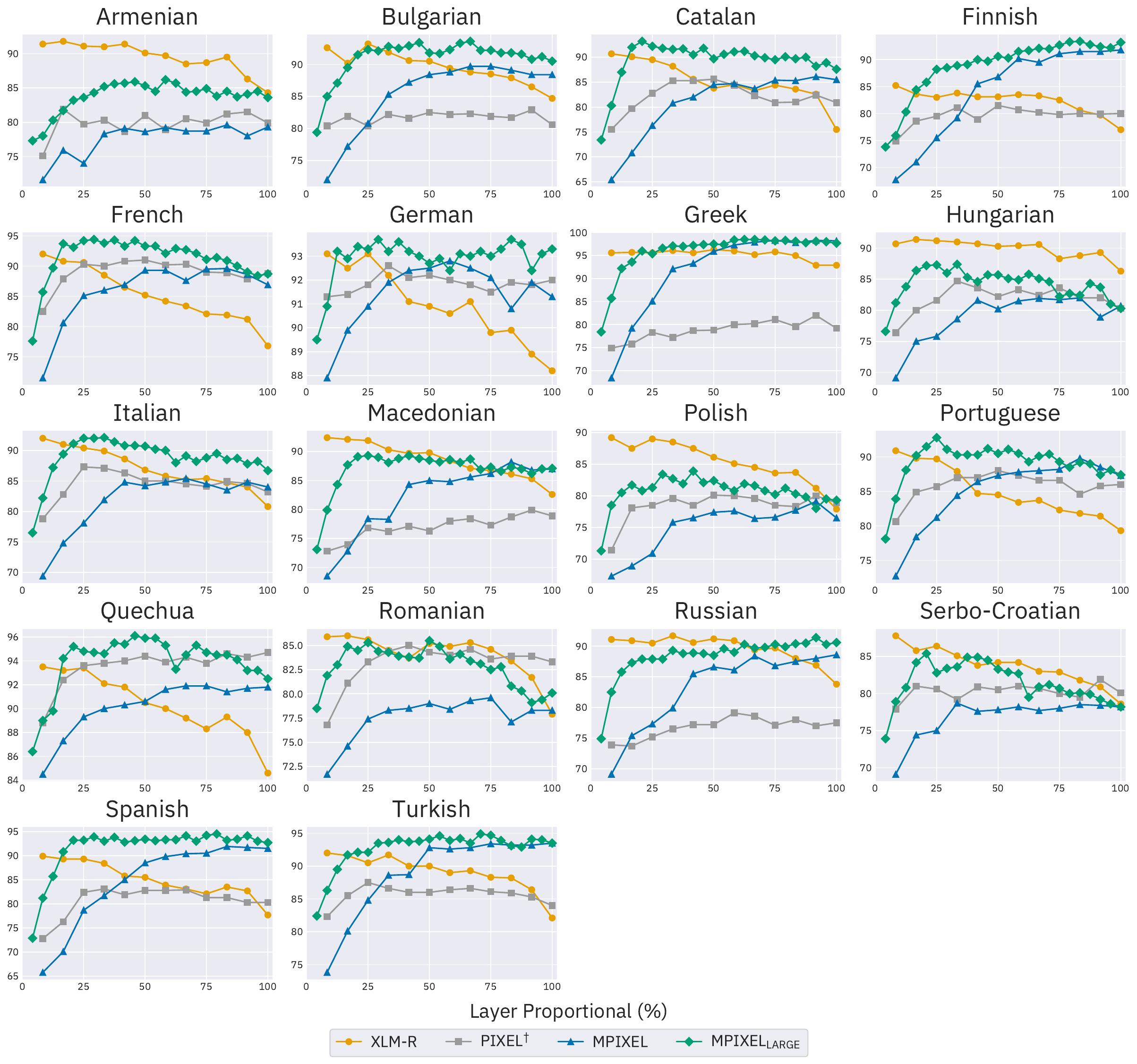}
  \caption{\textsc{linspector} probing results for the \textit{Tense} task. Each panel shows linear probe accuracy per layer, with the x-axis giving the layer position relative to the model depth.}
  \label{fig:linspector-tense}
\end{figure*}

\begin{figure*}[!htbp]
  \centering
  \includegraphics[width=\textwidth]{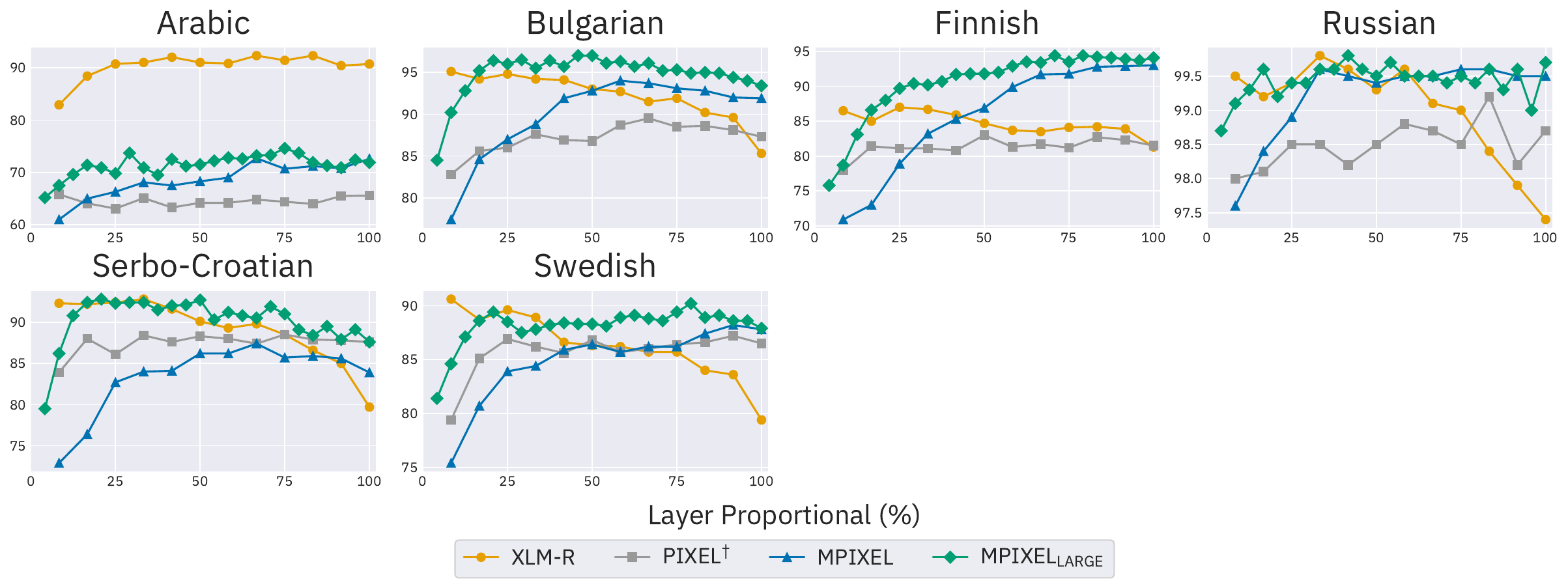}
  \caption{\textsc{linspector} probing results for the \textit{Voice} task. Each panel shows linear probe accuracy per layer, with the x-axis giving the layer position relative to the model depth.}
  \label{fig:linspector-voice}
\end{figure*}

\begin{figure}[!htbp]
  \centering
  \includegraphics[width=0.25\textwidth]{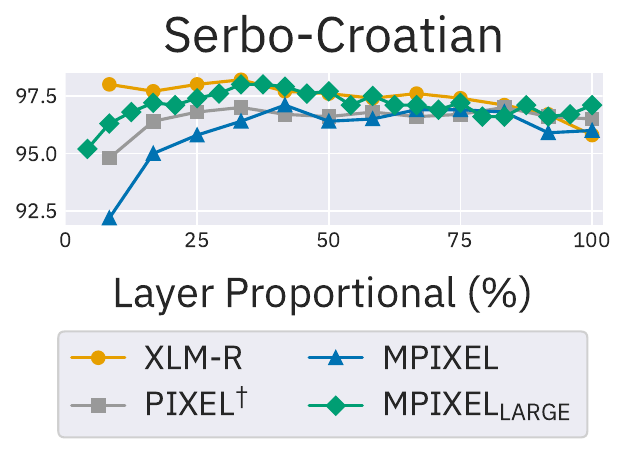}
  \caption{\textsc{linspector} probing results for the \textit{Comparison} task. Each panel shows linear probe accuracy per layer, with the x-axis giving the layer position relative to the model depth.}
  \label{fig:linspector-comparison}
\end{figure}

\begin{figure}[!htbp]
  \centering
  \includegraphics[width=0.25\textwidth]{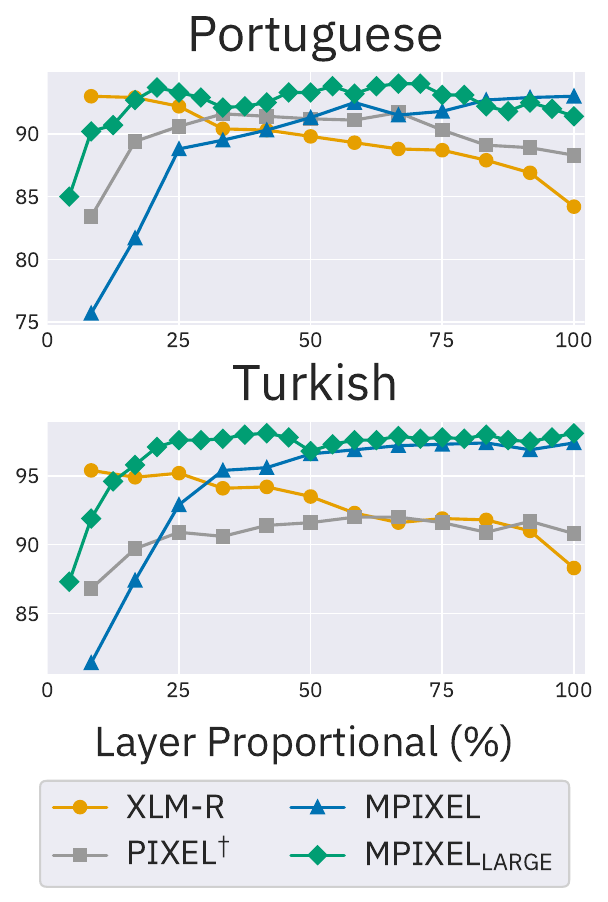}
  \caption{\textsc{linspector} probing results for the \textit{Polarity} task. Each panel shows linear probe accuracy per layer, with the x-axis giving the layer position relative to the model depth.}
  \label{fig:linspector-polarity}
\end{figure}

This appendix subsection contains the complete set of results for the word-level probing analyses on \textsc{linspector}~\citep{sahin-etal-2020-linspector}.
For these analyses, we use the entire set of languages and tasks included within \textsc{linspector}.
The \textsc{linspector} includes the following languages: Arabic, Armenian, Basque, Bulgarian, Catalan, Czech, Danish, Dutch, English, Estonian, Finnish,  French, German, Hungarian, Italian, Macedonian, Greek, Polish, Portuguese, Quechua, Romanian, Russian, Serbian, Serbo-Crotian, Spanish, Swedish, Turkish, Vietnamese.
The \textsc{linspector} task list consists of \textit{Case} (Fig.~\ref{fig:linspector-case}), \textit{Gender} (Fig.~\ref{fig:linspector-gender}), \textit{Mood} (Fig.~\ref{fig:linspector-mood}), \textit{Number} (Fig.~\ref{fig:linspector-number}), \textit{OddFeat} (Fig.~\ref{fig:linspector-oddfeat}), \textit{POS} (Fig.~\ref{fig:linspector-pos}), \textit{Person} (Fig.~\ref{fig:linspector-person}), \textit{Possession} (Fig.~\ref{fig:linspector-possession}), \textit{Pseudo} (Fig.~\ref{fig:linspector-pseudo}), \textit{SameFeat} (Fig.~\ref{fig:linspector-samefeat}), \textit{TagCount} (Fig.~\ref{fig:linspector-tagcount}), \textit{Tense} (Fig.~\ref{fig:linspector-tense}), \textit{Voice} (Fig.~\ref{fig:linspector-voice}), \textit{Comparison} (Fig.~\ref{fig:linspector-comparison}) and \textit{Polarity} (Fig.~\ref{fig:linspector-polarity}).

\onecolumn
\clearpage
{\setlength{\tabcolsep}{2.5pt}\setlength{\LTcapwidth}{\textwidth}\setlength{\LTleft}{0pt}\setlength{\LTright}{0pt}\fontsize{9}{11}\selectfont
\begin{longtable}[!ht]{@{\extracolsep{\fill}\hspace{\tabcolsep}}llccccccccc}
\toprule
\multirow{2}{*}{Language} & \multirow{2}{*}{Script} & \canines & \mbert & \xlmr & \pixelours & \mpixel[4s] & \mpixel[8l] & \mpixel[8s] & \mpixel & \mpixel \\
 & & 127M & 179M & 270M & 86M & 86M & 86M & 86M & 86M & 307M \\
\endfirsthead
\toprule
\multirow{2}{*}{Language} & \multirow{2}{*}{Script} & \canines & \mbert & \xlmr & \pixelours & \mpixel[4s] & \mpixel[8l] & \mpixel[8s] & \mpixel & \mpixel \\
 & & 127M & 179M & 270M & 86M & 86M & 86M & 86M & 86M & 307M \\
\midrule
\endhead
\midrule
\multicolumn{11}{r}{\textit{Continued on next page}} \\
\endfoot
\bottomrule
\addlinespace[\abovecaptionskip]
\caption{SIB-200 per-language test F1. Scores are averaged over the 3 runs with the best validation F1. \canines scores given without a standard deviation are imported from \citet{kesen-etal-2025-multilingual}. \xlmr results are taken from \citet{adelani-etal-2024-sib}.}\label{tab:sib} \\
\endlastfoot
\midrule
\multicolumn{11}{l}{\textbf{Pretraining Languages}} \\
\midrule
\quad\textsc{arb} & Arab & 69.2{\scriptsize$\pm$4.0} & \underline{84.1}{\scriptsize$\pm$2.3} & \textbf{88.2}{\scriptsize$\pm$0.7} & 42.7{\scriptsize$\pm$3.2} & 48.2{\scriptsize$\pm$1.6} & 47.6{\scriptsize$\pm$4.5} & 73.8{\scriptsize$\pm$3.1} & 75.3{\scriptsize$\pm$2.6} & 77.5{\scriptsize$\pm$1.9} \\
\quad\textsc{ben} & Beng & 54.5\hphantom{{\scriptsize$\pm$0.0}} & 78.5{\scriptsize$\pm$1.6} & \textbf{83.1}{\scriptsize$\pm$1.2} & 48.6{\scriptsize$\pm$0.4} & 58.4{\scriptsize$\pm$1.9} & 55.7{\scriptsize$\pm$1.2} & 58.0{\scriptsize$\pm$3.1} & 78.2{\scriptsize$\pm$2.3} & \underline{79.2}{\scriptsize$\pm$1.2} \\
\quad\textsc{dan} & Latn & 72.7{\scriptsize$\pm$1.8} & 83.4{\scriptsize$\pm$0.6} & \textbf{88.0}{\scriptsize$\pm$0.7} & 67.3{\scriptsize$\pm$0.6} & 60.8{\scriptsize$\pm$1.4} & 81.4{\scriptsize$\pm$2.0} & 61.9{\scriptsize$\pm$1.4} & 82.7{\scriptsize$\pm$0.8} & \underline{84.1}{\scriptsize$\pm$0.6} \\
\quad\textsc{ell} & Grek & 65.3\hphantom{{\scriptsize$\pm$0.0}} & 74.0{\scriptsize$\pm$1.8} & \textbf{85.4}{\scriptsize$\pm$1.3} & 42.0{\scriptsize$\pm$1.3} & 49.6{\scriptsize$\pm$0.3} & 50.0{\scriptsize$\pm$1.5} & \underline{84.3}{\scriptsize$\pm$1.7} & 82.3{\scriptsize$\pm$1.8} & 81.5{\scriptsize$\pm$1.2} \\
\quad\textsc{eng} & Latn & 78.4\hphantom{{\scriptsize$\pm$0.0}} & 87.7{\scriptsize$\pm$1.9} & \textbf{89.9}{\scriptsize$\pm$0.6} & \underline{88.6}{\scriptsize$\pm$1.4} & 86.2{\scriptsize$\pm$1.9} & 82.5{\scriptsize$\pm$2.9} & 86.1{\scriptsize$\pm$1.2} & 81.6{\scriptsize$\pm$2.6} & 86.8{\scriptsize$\pm$1.6} \\
\quad\textsc{fin} & Latn & 61.6\hphantom{{\scriptsize$\pm$0.0}} & 78.2{\scriptsize$\pm$3.6} & \textbf{89.0}{\scriptsize$\pm$1.1} & 66.4{\scriptsize$\pm$1.5} & 61.3{\scriptsize$\pm$2.3} & 82.2{\scriptsize$\pm$1.5} & 64.6{\scriptsize$\pm$1.3} & 82.9{\scriptsize$\pm$2.6} & \underline{84.7}{\scriptsize$\pm$0.8} \\
\end{longtable}}
\twocolumn

\begin{table*}[!htbp]
\centering
\resizebox{\textwidth}{!}{%
\begin{tabular}{lccccccccc}
\toprule
 & \canines & \mbert & \xlmr & \pixelours & \mpixel[4s] & \mpixel[8l] & \mpixel[8s] & \mpixel & \mpixel \\
 & 127M & 179M & 270M & 86M & 86M & 86M & 86M & 86M & 307M \\
\midrule
\multicolumn{10}{l}{\textbf{Pretraining Languages}} \\
\midrule
\quad\textsc{eng} & 67.1\hphantom{{\scriptsize$\pm$0.0}} & \underline{81.5}{\scriptsize$\pm$1.1} & \textbf{83.2}{\scriptsize$\pm$0.7} & 81.1{\scriptsize$\pm$0.8} & 78.9{\scriptsize$\pm$0.2} & 77.0{\scriptsize$\pm$0.5} & 75.5{\scriptsize$\pm$0.8} & 74.9{\scriptsize$\pm$1.8} & 80.9{\scriptsize$\pm$0.7} \\
\quad\textsc{dan} & 70.5{\scriptsize$\pm$0.4} & \underline{86.5}{\scriptsize$\pm$0.2} & \textbf{86.6}{\scriptsize$\pm$0.6} & 76.3{\scriptsize$\pm$1.1} & 74.1{\scriptsize$\pm$0.3} & 80.7{\scriptsize$\pm$1.6} & 74.3{\scriptsize$\pm$0.1} & 78.4{\scriptsize$\pm$2.1} & 85.8{\scriptsize$\pm$0.2} \\
\quad\textsc{ell} & 56.1{\scriptsize$\pm$1.5} & 78.8{\scriptsize$\pm$0.4} & \textbf{81.1}{\scriptsize$\pm$0.6} & 60.9{\scriptsize$\pm$0.8} & 62.5{\scriptsize$\pm$0.8} & 62.9{\scriptsize$\pm$1.4} & 75.5{\scriptsize$\pm$1.0} & 70.0{\scriptsize$\pm$0.5} & \underline{79.8}{\scriptsize$\pm$0.2} \\
\quad\textsc{heb} & 42.0{\scriptsize$\pm$1.5} & \underline{73.4}{\scriptsize$\pm$0.7} & \textbf{78.4}{\scriptsize$\pm$0.7} & 57.1{\scriptsize$\pm$1.0} & 59.5{\scriptsize$\pm$1.7} & 58.0{\scriptsize$\pm$1.0} & 58.3{\scriptsize$\pm$1.4} & 68.2{\scriptsize$\pm$1.0} & 72.9{\scriptsize$\pm$0.6} \\
\quad\textsc{zho} & 75.8\hphantom{{\scriptsize$\pm$0.0}} & \textbf{87.5}{\scriptsize$\pm$1.1} & \underline{86.4}{\scriptsize$\pm$0.3} & 64.4{\scriptsize$\pm$0.5} & 81.0{\scriptsize$\pm$0.3} & 80.5{\scriptsize$\pm$0.4} & 79.8{\scriptsize$\pm$0.7} & 77.7{\scriptsize$\pm$0.6} & 81.7{\scriptsize$\pm$0.1} \\
\bottomrule
\end{tabular}}%
\caption{NER per-language test F1. Validation split F1 is reported for \textsc{kor}, as it does not include a test set. Scores are averaged over the 3 runs with the best validation F1. CANINE-S scores given without a standard deviation are imported from \citet{kesen-etal-2025-multilingual}.}
\label{tab:ner}
\end{table*}

\begin{table*}[!htbp]
\centering
\resizebox{\textwidth}{!}{%
\begin{tabular}{lcccccccccccccccccc}
\toprule
Model & $\theta$ & \#L & \textsc{arb} & \textsc{bul} & \textsc{deu} & \textsc{ell} & \textsc{eng} & \textsc{spa} & \textsc{fra} & \textsc{hin} & \textsc{rus} & \textsc{swa} & \textsc{tha} & \textsc{tur} & \textsc{urd} & \textsc{vie} & \textsc{zho} & Avg \\
\midrule
\canines & 127M & 104 & 69.7\hphantom{{\scriptsize$\pm$0.0}} & 75.0\hphantom{{\scriptsize$\pm$0.0}} & 74.9\hphantom{{\scriptsize$\pm$0.0}} & 72.5\hphantom{{\scriptsize$\pm$0.0}} & 79.1\hphantom{{\scriptsize$\pm$0.0}} & 76.3\hphantom{{\scriptsize$\pm$0.0}} & 75.3\hphantom{{\scriptsize$\pm$0.0}} & 65.2\hphantom{{\scriptsize$\pm$0.0}} & 73.0\hphantom{{\scriptsize$\pm$0.0}} & 65.0\hphantom{{\scriptsize$\pm$0.0}} & 62.3\hphantom{{\scriptsize$\pm$0.0}} & 68.9\hphantom{{\scriptsize$\pm$0.0}} & 64.1\hphantom{{\scriptsize$\pm$0.0}} & 71.3\hphantom{{\scriptsize$\pm$0.0}} & 65.6\hphantom{{\scriptsize$\pm$0.0}} & 70.5\hphantom{{\scriptsize$\pm$0.0}} \\
\mbert & 179M & 104 & 73.2\hphantom{{\scriptsize$\pm$0.0}} & \underline{77.7}\hphantom{{\scriptsize$\pm$0.0}} & \underline{77.5}\hphantom{{\scriptsize$\pm$0.0}} & 75.7\hphantom{{\scriptsize$\pm$0.0}} & 83.5\hphantom{{\scriptsize$\pm$0.0}} & 79.8\hphantom{{\scriptsize$\pm$0.0}} & \underline{78.6}\hphantom{{\scriptsize$\pm$0.0}} & 70.1\hphantom{{\scriptsize$\pm$0.0}} & \underline{76.4}\hphantom{{\scriptsize$\pm$0.0}} & \underline{68.1}\hphantom{{\scriptsize$\pm$0.0}} & 67.2\hphantom{{\scriptsize$\pm$0.0}} & 73.8\hphantom{{\scriptsize$\pm$0.0}} & \underline{64.4}\hphantom{{\scriptsize$\pm$0.0}} & 76.5\hphantom{{\scriptsize$\pm$0.0}} & 77.9\hphantom{{\scriptsize$\pm$0.0}} & 74.7\hphantom{{\scriptsize$\pm$0.0}} \\
\bottomrule
\end{tabular}}%
\caption{XNLI per-language test accuracy. Scores are averaged over the 3 runs with the best validation accuracy. \xlmr scores are imported from \citet{conneau-etal-2020-unsupervised}, and \canines and \mbert scores from \citet{sun-etal-2023-multi}.}
\label{tab:xnli-test}
\end{table*}

\begin{table*}[!htbp]
\centering
\resizebox{\textwidth}{!}{%
\begin{tabular}{lcccccccccccc}
\toprule
\multirow{2}{*}{Model} & \multirow{2}{*}{$\theta$} & \multirow{2}{*}{\#L} & MNLI-M/MM & QQP & QNLI & SST-2 & CoLA & STS-B & MRPC & RTE & WNLI & \multirow{2}{*}{Avg} \\
 &  &  & 393k & 364k & 105k & 67k & 8.6k & 5.8k & 3.7k & 2.5k & 635 &  \\
\midrule
\canines & 127M & 104 & 77.3{\scriptsize$\pm$0.1} / 78.0{\scriptsize$\pm$0.1} & 85.3{\scriptsize$\pm$0.2} & 88.8{\scriptsize$\pm$0.2} & 88.0{\scriptsize$\pm$0.0} & 17.4{\scriptsize$\pm$0.6} & 85.1{\scriptsize$\pm$0.0} & 88.3{\scriptsize$\pm$0.3} & 65.0{\scriptsize$\pm$0.6} & 56.8{\scriptsize$\pm$0.7} & 73.0\hphantom{{\scriptsize$\pm$0.0}} \\
\mbert & 179M & 104 & 82.3{\scriptsize$\pm$0.1} / 82.8{\scriptsize$\pm$0.1} & 87.3{\scriptsize$\pm$0.0} & \textbf{91.1}{\scriptsize$\pm$0.2} & 91.1{\scriptsize$\pm$0.2} & 46.1{\scriptsize$\pm$1.1} & \textbf{89.2}{\scriptsize$\pm$0.0} & 90.8{\scriptsize$\pm$0.1} & \textbf{76.7}{\scriptsize$\pm$0.9} & 58.7{\scriptsize$\pm$1.8} & 79.6\hphantom{{\scriptsize$\pm$0.0}} \\
\xlmr & 270M & 100 & \textbf{84.7}{\scriptsize$\pm$0.1} / \textbf{84.7}{\scriptsize$\pm$0.1} & \underline{88.0}{\scriptsize$\pm$0.1} & \underline{90.6}{\scriptsize$\pm$0.0} & \textbf{93.1}{\scriptsize$\pm$0.1} & \textbf{60.3}{\scriptsize$\pm$0.2} & \underline{88.8}{\scriptsize$\pm$0.1} & \textbf{92.0}{\scriptsize$\pm$0.1} & \underline{69.0}{\scriptsize$\pm$0.3} & 56.3{\scriptsize$\pm$0.0} & \textbf{80.8}\hphantom{{\scriptsize$\pm$0.0}} \\
\midrule
\pixelours & 86M & 1 & \underline{83.2}{\scriptsize$\pm$0.1} / \underline{83.6}{\scriptsize$\pm$0.3} & 87.5{\scriptsize$\pm$0.2} & 89.6{\scriptsize$\pm$0.1} & 91.9{\scriptsize$\pm$0.1} & 49.4{\scriptsize$\pm$0.4} & 88.5{\scriptsize$\pm$0.0} & \underline{91.4}{\scriptsize$\pm$0.2} & 65.6{\scriptsize$\pm$0.5} & 56.8{\scriptsize$\pm$1.3} & 78.7\hphantom{{\scriptsize$\pm$0.0}} \\
\bottomrule
\end{tabular}}%
\caption{GLUE test results. Scores are averaged over the 3 runs with the best validation score.}
\label{tab:glue}
\end{table*}

\begin{table*}[!htbp]
\centering
\resizebox{\textwidth}{!}{%
{\setlength{\tabcolsep}{4pt}\fontsize{10}{12}\selectfont
\begin{tabular}{@{}llccccccccc@{}}
\toprule
Language & Script & \canines & \mbert & \xlmr & \pixelours & \mpixel[4s] & \mpixel[8l] & \mpixel[8s] & \mpixel & \mpixel \\
$|\theta|$ & & 127M & 179M & 270M & 86M & 86M & 86M & 86M & 86M & 307M \\
\midrule
\multicolumn{11}{l}{\textbf{Pretraining Languages}} \\
\midrule
\quad\textsc{arb} & Arab & 77.5\hphantom{{\scriptsize$\pm$0.0}} & \underline{85.5}{\scriptsize$\pm$0.1} & \textbf{86.6}{\scriptsize$\pm$0.1} & 80.0{\scriptsize$\pm$0.0} & 79.4{\scriptsize$\pm$0.1} & 79.8{\scriptsize$\pm$0.0} & 82.8{\scriptsize$\pm$0.2} & 82.2{\scriptsize$\pm$0.1} & 83.2{\scriptsize$\pm$0.3} \\
\quad\textsc{zho} & Hant & 75.1\hphantom{{\scriptsize$\pm$0.0}} & \underline{85.4}{\scriptsize$\pm$0.1} & \textbf{86.5}{\scriptsize$\pm$0.3} & 71.2{\scriptsize$\pm$0.2} & 83.0{\scriptsize$\pm$0.4} & 82.6{\scriptsize$\pm$0.5} & 82.4{\scriptsize$\pm$0.2} & 81.6{\scriptsize$\pm$0.3} & 85.0{\scriptsize$\pm$0.2} \\
\quad\textsc{dan} & Latn & 68.0{\scriptsize$\pm$0.7} & 87.5{\scriptsize$\pm$0.1} & \textbf{89.8}{\scriptsize$\pm$0.4} & 80.3{\scriptsize$\pm$0.1} & 80.2{\scriptsize$\pm$0.3} & 86.6{\scriptsize$\pm$0.1} & 80.3{\scriptsize$\pm$0.2} & 85.7{\scriptsize$\pm$0.3} & \underline{88.2}{\scriptsize$\pm$0.1} \\
\quad\textsc{eng} & Latn & 82.4\hphantom{{\scriptsize$\pm$0.0}} & 91.8{\scriptsize$\pm$0.2} & \textbf{93.2}{\scriptsize$\pm$0.1} & \underline{92.5}{\scriptsize$\pm$0.1} & 89.3{\scriptsize$\pm$0.2} & 89.2{\scriptsize$\pm$0.1} & 88.7{\scriptsize$\pm$0.1} & 88.7{\scriptsize$\pm$0.2} & 90.5{\scriptsize$\pm$0.1} \\
\quad\textsc{fin} & Latn & 72.0{\scriptsize$\pm$0.7} & 87.6{\scriptsize$\pm$0.2} & 91.4{\scriptsize$\pm$0.1} & 80.4{\scriptsize$\pm$0.1} & 82.0{\scriptsize$\pm$0.2} & \underline{91.7}{\scriptsize$\pm$0.1} & 82.7{\scriptsize$\pm$0.2} & 90.6{\scriptsize$\pm$0.1} & \textbf{92.3}{\scriptsize$\pm$0.4} \\
\bottomrule
\end{tabular}}}%
\caption{UDP per-language test LAS. Scores are averaged over the 3 runs with the best validation LAS. \canines scores given without a standard deviation are imported from \citet{kesen-etal-2025-multilingual}.}
\label{tab:udp}
\end{table*}

 \clearpage

\subsection{ABX More Ablations}
\label{sec:appendix_abx}

\subsubsection{Generalization to unseen languages.}
\label{sec:appendix_abx_unseen}

To test whether scaling multilinguality affects seen and unseen languages differently,
we compare (seen, seen), (seen, unseen), and (unseen, unseen) language pairs across our \pixel 
models with increasing language coverage. 

\begin{figure}
    \centering
    \includegraphics[width=0.5\textwidth]{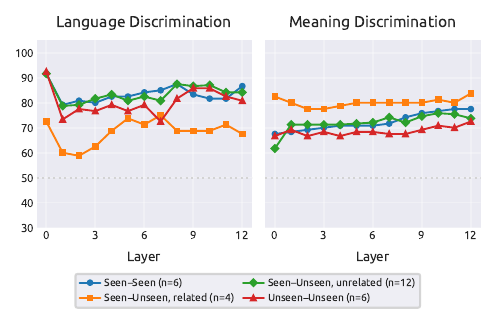}
        \includegraphics[width=0.5\textwidth]{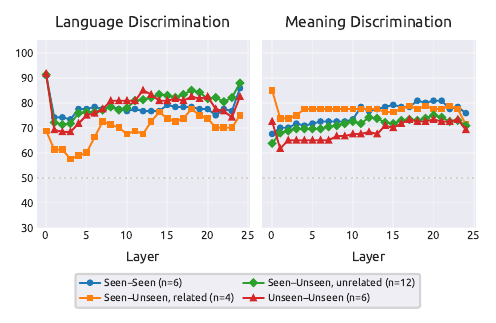}
    \caption{ABX probes, \pixelmsixteenbase (above) and \pixelmsixteenlarge (below). Languages seen and unseen during pre-training, in different combinations. }
    \label{fig:abx_seen_unseen}
\end{figure}

We test whether alignment extends to languages held out of pretraining by evaluating \pixelmsixteenbase on four unseen languages (Assamese, Icelandic, Yiddish, North Azerbaijani), each chosen to share its script (and, in three of four cases, its language family) with one of four seen languages (Bengali, Danish, Hebrew, Turkish).
Splitting the language pairs into seen--seen, seen--unseen (related vs.\ unrelated), and unseen--unseen groups (Figure~\ref{fig:abx_seen_unseen}), the related seen--unseen pairs stand out: they have the highest, flattest MD and the lowest LD at every depth. %
Unseen--unseen pairs have the lowest MD scores and an unstable LD trajectory. The difference between seen--seen pairs and unrelated seen--unseen pairs is small, indicating that joint inclusion in pre-training does not increase alignment. 

\subsubsection{Same- versus cross-script pairs.}
\label{sec:appendix_abx_script}

Extending the language set from the \pixelmfour languages to all 16 languages lowers LD and raises MD scores. For \pixelmsixteenbase, embedding-layer LD falls from $1.00$ on the four \pixelmfour languages to $0.95$ on all 16. The reason is the changed pair composition: the 16-language set contains many same-script pairs, which lack the visual cues that make cross-script pairs trivially language-discriminable and are easier to align in meaning. 

\begin{figure}
    \centering
    \includegraphics[width=0.5\textwidth]{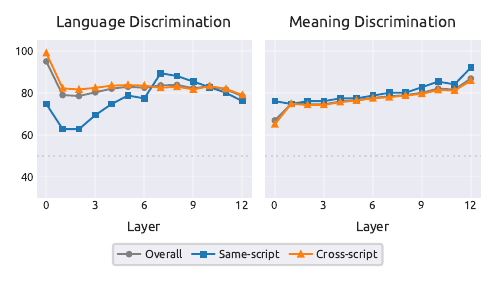}
        \includegraphics[width=0.5\textwidth]{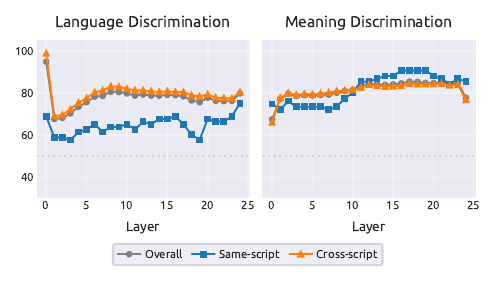}
    \caption{ABX probes, \pixelmsixteenbase (above) and \pixelmsixteenlarge (below), \pixelmsixteenbase pre-training languages. Pairs of languages that share script (blue) versus pairs of languages that have different scripts (orange). }
    \label{fig:abx_same_vs_cross_script}
\end{figure}

Splitting pairs by whether the two languages share a script (same-script vs.\ cross-script) confirms this; results are in Figure~\ref{fig:abx_same_vs_cross_script}. The clearest effect is at the embedding layer of LD: cross-script pairs are near-trivially discriminable%
, whereas same-script pairs start far lower. 
Cross-script LD decreases substantially already in the first layer. %
The two model sizes differ in how the script signal is then
handled. 
In \pixelmsixteenbase, same-script LD recovers and even exceeds cross-script LD in the middle layers.
MD is slightly higher within the same script at every depth and climbs monotonically for both groups. 
In \pixelmsixteenlarge however, same-script LD stays well below cross-script LD throughout and is less stable. 

The same-script MD trajectory is where the two models diverge most.
For \pixelmsixteenbase, MD is only mildly affected by script sharing: the curves differ mostly in the first layer, where the same-script MD is notably higher. 
In \pixelmsixteenlarge, the MD scores for same-script pairs are much more dynamic (and peak higher) than the ones of cross-script pairs. 

\subsection{Script ID Probes}
\label{sec:appendix_script_probes}

To test how explicitly script identity is encoded across layers, we train linear probes on each layer's representations to classify text into the 10 scripts seen during \pixelmsixteenbase pretraining. 

\paragraph{Setup} We train linear probing classifiers on the mean-pooled encoder representations at each of the 13 layers. Each probe is a multinomial logistic regression (L2 regularization, $C{=}1.0$) trained to predict script identity from the BOUQuET dataset \citep{andrews-etal-2025-bouquet}. 
We evaluate in two settings: 
\emph{iid}, on unseen sentences from the 16 training languages, and
\emph{ood}, on 32 unseen languages whose scripts (Arabic, Bengali, Cyrillic, Devanagari, Han, Hebrew, Latin) the probe has seen. 

\paragraph{\pixel models} 
In the iid setting, script identity is recovered almost perfectly at every layer in all base models. %
The models diverge in the ood setting:
\pixelmsixteenbase and \pixelmeightb retain near-perfect accuracy across all layers, whereas \pixelmfour and \pixelmeighta decline in the final layers%
; a clear, though not catastrophic, drop from the perfect accuracy of the early layers and of the higher-coverage models. %
The high results of the \pixelours model however show that the model trained on only one script also behaves robustly. 

Notably, the late-layer errors of \mpixel[4s] and \mpixel[8l] are largely Cyrillic misclassified as Latin, even though every \mpixel variant has exactly one Cyrillic-script pretraining language (Ukrainian). %

\paragraph{Visual versus token-based inputs}
Compared to XLM-R, we see that \pixel models generalize more reliably to out-of-distribution languages, indicating a difference of the visual versus subword token-based processing for unseen languages..
\begin{figure}
    \centering
    \includegraphics[width=0.5\textwidth]{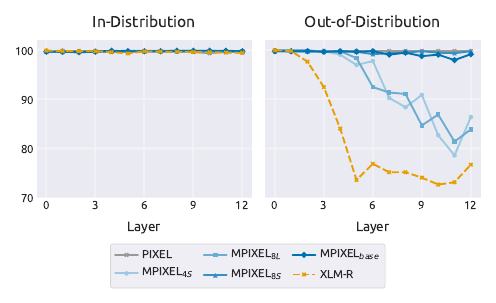}
    \caption{Layer-wise accuracy of linear script-identification probes trained on the representations of each base model. \emph{Left (iid):} unseen sentences from the 16 training languages. \emph{Right (ood):} 32 unseen languages whose scripts were seen during training. }
    \label{fig:script_id_probe}
\end{figure}

\clearpage

\end{document}